\documentclass[10pt,journal,compsoc]{IEEEtran}

\ifCLASSOPTIONcompsoc
  \usepackage[nocompress]{cite}
\else
  \usepackage{cite}
\fi

\usepackage[colorlinks,
            linkcolor=blue,
            anchorcolor=blue,
            citecolor=green]{hyperref}
\usepackage{amsmath}
\usepackage{subfigure}
\usepackage{caption}
\usepackage{algorithm}
\usepackage{algpseudocode}
\usepackage{setspace}
\usepackage{color,hyperref}

\usepackage{graphicx}
\usepackage{amsmath,amssymb} 
\usepackage{epsfig}
\usepackage{amsmath}
\usepackage{amssymb}
\usepackage{colortbl}
\usepackage{multirow}
\usepackage{pifont}
\usepackage{url}
\usepackage{soul}
\usepackage[table ]{ xcolor}
\soulregister\cite7
\soulregister\ref7

\newcolumntype{I}{!{\vrule width 1.2pt}}
\newlength\savedwidth
\newcommand\whline{\noalign{\global\savedwidth\arrayrulewidth
		\global\arrayrulewidth 1.25pt}%
	\hline
	\noalign{\global\arrayrulewidth\savedwidth}}

\usepackage{stfloats}
\begin{document}
%
\title{LDC-Net: A Unified Framework for Localization, Detection and Counting in Dense Crowds}

%

\author{Qi~Wang,~\IEEEmembership{Senior~Member,~IEEE,}
        Tao~Han, Junyu~Gao,
        Yuan~Yuan,~\IEEEmembership{~Senior Member,~IEEE,}
        Xuelong~Li,~\IEEEmembership{~Fellow,~IEEE}

\thanks{
The authors are with the School of Artificial Intelligence, Optics and Electronics (iOPEN), Northwestern Polytechnical University, Xi'an 710072, P.R. China. (e-mail: crabwq@gmail.com; hantao10200@mail.nwpu.edu.cn; gjy3035@gmail.com; y.yuan1.ieee@gmail.com; li@nwpu.edu.cn).
}}

\IEEEtitleabstractindextext{%
\begin{abstract}
The rapid development in visual crowd analysis shows a trend to count people by positioning or even detecting, rather than simply summing a density map. It also enlightens us back to the essence of the field, detection to count, which can give more abundant crowd information and has more practical applications. However, some recent work on crowd localization and detection has two limitations: 1) The typical detection methods can not handle the dense crowds and a large variation in scale; 2) The density map heuristic methods suffer from performance deficiency in position and box prediction, especially in high density or large-size crowds. In this paper, we devise a tailored baseline for dense crowds location, detection, and counting from a new perspective, named as LDC-Net for convenience, which has the following features: 1) A strong but minimalist paradigm to detect objects by only predicting a location map and a size map, which endows an ability to detect in a scene with any capacity ($0 \sim 10,000+$ persons); 2) Excellent cross-scale ability in facing a large variation, such as the head ranging in $0 \sim 100,000+$ pixels; 3) Achieve superior performance in location and box prediction tasks, as well as a competitive counting performance compared with the density-based methods. Finally, the source code and pre-trained models will be released.
\end{abstract}

}

\maketitle

\IEEEdisplaynontitleabstractindextext

%
\IEEEpeerreviewmaketitle

\IEEEraisesectionheading{\section{Introduction}\label{sec:introduction}}

\IEEEPARstart{A}{t} present, crowd analysis in dense scenes mainly concentrates on the heads, which usually has less occlusion compared with the body and provides more information for digging. The past decade has shown the rapid development of crowds analysis on heads in dense scenarios, such as crowd detection and crowd counting \cite{chan2008privacy, li2018csrnet,wan2019residual, ma2019bayesian, wan2020modeling}. The former makes an instance-level prediction with a detector supervised by the box-level annotation. The latter learns from the point annotations to output the image-level counting number by summing the predicted density map. However, density-based crowd counting shows the saturated property in recent years. With the emergence of large-scale and densely box-level annotated datasets (NWPU-Crowd \cite{gao2020nwpu} and JHU-CROWD \cite{sindagi2019pushing,sindagi2020jhu}), a new branch extends the field and has attracted a lot of attention from researchers, termed crowd localization. From the perspective of social needs, the research on these basic tasks contributes to public safety management. Besides, they are also fundamental to the high-level crowd analysis tasks \cite{kang2014fully,li2017multiview}, such as group detection in crowded scenarios, crowd tracking, etc.

\begin{figure}[t]
	\centering
	\includegraphics[width=0.48\textwidth]{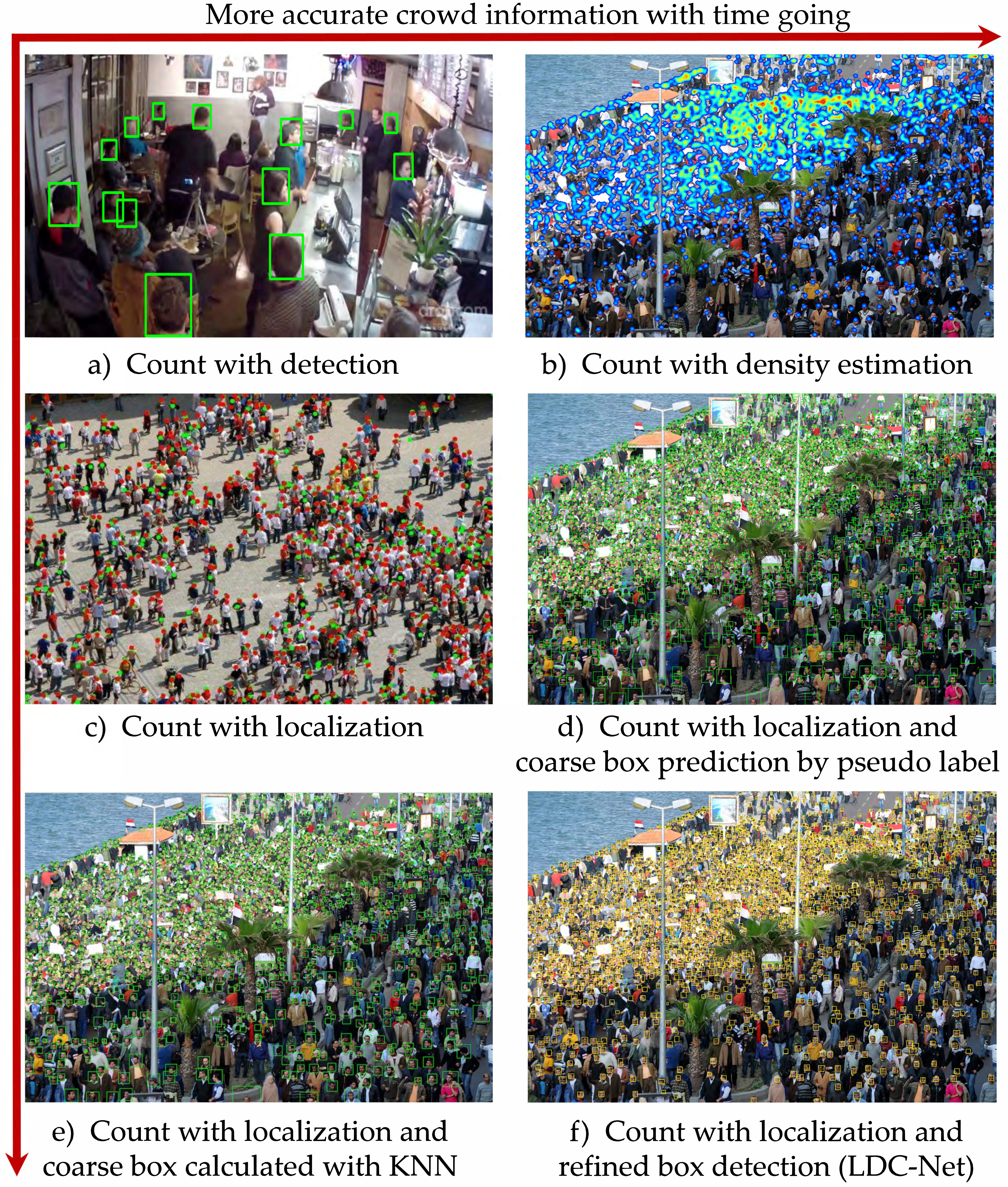}
	\caption{Trends of crowd counting. More information in the dense scene is being given: count number $\rightarrow$ individual location $\rightarrow$ box-level predictions. a), c), d) and e) come from E2EPD \cite{stewart2016end}, RAZ\_loc \cite{liu2019recurrent}, LSC-CNN \cite{sam2020locate}, and FIDTM \cite{liang2021focal}, respectively.}
	\label{fig:introduction}
\end{figure}

For a long time, researchers treat the three tasks independently. Crowd detection is usually conducted in the low-density scene (e.g., people detection \cite{stewart2016end} in Fig .\ref{fig:introduction} a), which inherits and develops the methods from the object detection field. And the extensive research show it has achieved outstanding performance in some sparse scenes. However, such methods have natural defects in detecting dense crowds (e.g., 1,000+ people in an image) and small heads (below 100 pixels). On the other hand, the density estimation crowd counting methods, depicted in Fig. \ref{fig:introduction} b), can handle the extremely dense scene better. Such approaches, however, do not take account of the individuals and cannot provide the position or scale information of the heads. As for crowd localization, it is an under-explored domain. And a couple of methods that come up with the detection method and heuristic algorithms (extract local extreme points from density/segmentation-score maps as illustrated in Fig. \ref{fig:introduction} c). However, the localization performance is unsatisfactory because they fail at some small scale and congested scenes.

In this paper, we strive to treat the three tasks with a single framework by designing a well-performed detector in dense crowd to handle the tasks of locating and counting simultaneously. Generally, there are some challenges in building such a trinity framework: a) Individual diversity in the poses, viewpoints, and illumination variations; b) Large scale variations in dense scenes; c) The resolution reduction leads to missing head positions in dense areas; d) Extreme box sizes (beyond 100,000 pixels in NWPU-Crowd dataset) exceed the detection capability of the detector; e) The number of different box size takes long-tail distribution; f) Box level annotations are absent in some datasets.

Some methods \cite{liu2019point, sam2020locate, liang2021focal, wang2021self} try to tackle the above challenges by proposing a powerful detection framework. Among them, LDC-Net \cite{sam2020locate} designs an end-to-end detection-based dense crowd counting framework, which achieves superior positioning and counting performance. However, as a scheme that inherits from traditional detection methods, it inevitably has the following problems: 1) the model structure is complex, and the computational complexity is high. 2) As shown in Fig. \ref{fig:introduction} d), it does not achieve accurate size estimation with its predefined pseudo box labels. 3) It must set the threshold manually or implement a parameter search to select threshold, which restricts the cross-scene application of the model. 4) It has good counting performance but is weak in positioning and detection, especially in extremely dense crowds. Besides, FIDTM \cite{liang2021focal} generates the box prediction according to the KNN algorithm, which is a non-parameter method and fails in sparse scenes as Fig. \ref{fig:introduction} e) shown.

Different from the existing methods, this paper aims to bring a new insight for resolving detection, localization and counting in dense crowds. As we mentioned above, crowd analysis mainly focuses on the heads, and each head usually occupies a few pixels, even a tiny object. The detection would be easy if the heads are segmented into some independent areas. This task seems a bit similar to instance segmentation. However, it would fail to directly segment the head area because the neighboring heads may be connected in a dense crowds.  Besides, the high similarity of the head appearance will make the instance segmentation extremely difficult. Hence, we follow the IIM \cite{gao2020learning} to utilize Independent Instance Map for crowd localization.

In conclusion, this paper proposes a framework that locates, detects, and counts people in the dense scene with promising performance, LDC-Net for abbreviation. Specifically,  LDC-Net first converts the crowd scene into a confidence map with a convolutional neural network. Then, we segment it to a binary map with the binarization module, which adaptively predicts a pixel-level threshold map for every image to achieve optimal segmentation. Compared with the manual thresholds, it is more flexible, efficient, and intelligent. Finally, each blob in the binary map corresponds to an individual, the head positions and initial boxes are generated by detecting the connected components from the binary map. To detect the people in the dense scene, we devise a size prediction branch, which learns a size map, where the value in each head location represents its size. By integrating the location information, initial box, and size map, the elaborate bounding boxes are generated for objects (Fig. \ref{fig:introduction} f). Finally, the contributions of this work are concluded as:

\begin{enumerate}
\item[1)] \textbf{LDC-Net:} Propose a neat and novel end-to-end Localization, Detection, and Counting framework to cope with dense crowds, which explores a brand new approach to make the detection in ultrahigh density scene efficiently and accurately in comparison with the existing methods.

\item[2)] \textbf{Pixel-wise Binarization:} Design a stable threshold learner to adaptively generate the pixel-level thresholds according to the high-level semantic information, which embeds into the localization module and no longer requires extra supervised labels. The pixel-wise thresholds can catch more hard objects compared with the commonly fixed threshold.

\item[3)] \textbf{OUSR Loss:} Propose an Over and Under Segmentation Ratio objective function to optimize the LDC-Net, which refines the confidence map to has fewer conglutinant blobs in dense crowds and significantly improve the recall rate of object localization.

\item[4)] \textbf{Size Estimation:} Introduce a new object size modeling method, which uses one channel scale map to describe the object size in the dense crowds. It is a simplistic detection strategy but can get an accurate box prediction.

\end{enumerate}

The rest of this paper takes the form of four chapters. Section \ref{sec:Related Works} gives a comprehensive overview of the related work in terms of crowd detection, counting, and localization. Section \ref{sec:Proposed LDC-Net} concerns the methodology used in LDC-Net, which details the location, detection, and counting method. Besides, it also introduces the loss functions to supervise the training of LDC-Net. Section \ref{sec:Experiments} is about the experiments, which first introduces the datasets, training details, evaluation metrics. Then, an ablation study is conducted to discuss some proposed modules. Finally, many experiments are designed to comprehensively evaluate the localization, detection, and counting performance for LDC-Net in association with the previous methods. In the end, the conclusion of this paper is summarized in Section \ref{sec:Conclusion}.

\section{Related Works}
\label{sec:Related Works}

\subsection{Crowd Counting}
The state-of-the-art crowd counting methods are mostly concentrated on density map estimation in recent years, which integrates the density map as a count value. CNN-based methods \cite{zhang2016single,onoro2016towards,sam2017switching,bai2020adaptive} show its powerful capacity of feature extraction than hand-crafted features models \cite{idrees2013multi,liu2015bayesian}. Some methods \cite{shi2018crowd,shi2019revisiting,cheng2019learning,jiang2020attention,liu2020adaptive} work on network architectures or specific modules to regress pixel-wise or patch-wise density maps. In addition to density-map supervision, some methods \cite{ma2019bayesian,wang2020distribution,dong2020scale} directly exploit point-level labels to supervise counting models. Unfortunately, these methods only predict image-level counts or coarse local density. The position of each head is hard to obtain accurately.

\subsection{Crowd Localization}
\label{sec2:crowd_localizaiton}
Three typical methods focus on dense crowds localization. 1) As detailed in Sec. \ref{sec:crowd_detection}, detection-based methods use the center point in the predicted box as location.  2) Heuristic algorithms \cite{idrees2018composition,liu2019recurrent,gao2019domain} find the peak value position as the head location in the density or heat map. 3) Point-supervision methods \cite{laradji2018blobs,liu2019point,sam2020locate} directly use point-level labels or generate pseudo box labels to learn to locate the head position.

\textbf{Heuristic Crowd Localization.\,\,\,\,}
Benefiting from the high-resolution and fine density map, some methods try to find head locations from them. Idrees \emph{et al.} \cite{idrees2018composition} and Liu \emph{et al.} \cite{liu2019recurrent} propose a post-processing method by finding the peak point in a local region ($3 \times 3$ pixels) as the final head position. Gao \emph{et al.} \cite{gao2019domain} present a Gaussian-prior Reconstruction algorithm to recover the center of $15 \times 15$ Gaussian Kernel from predicted density maps. In addition to the density map, some works propose new heat maps to extract the local extreme point. FIDTM \cite{liang2021focal} propose a distance transform map to describe the probability of each pixel as a head location. D2CNet \cite{cheng2021decoupled} predicts the probability map, of which each pixel represents the probability of being an object. These methods can locate tiny objects well. However, it may also output multiple predictions for one large-scale head or dense regions because they rely on fixed window size.

\begin{figure*}[t]
	\centering
	\includegraphics[width=0.9\textwidth]{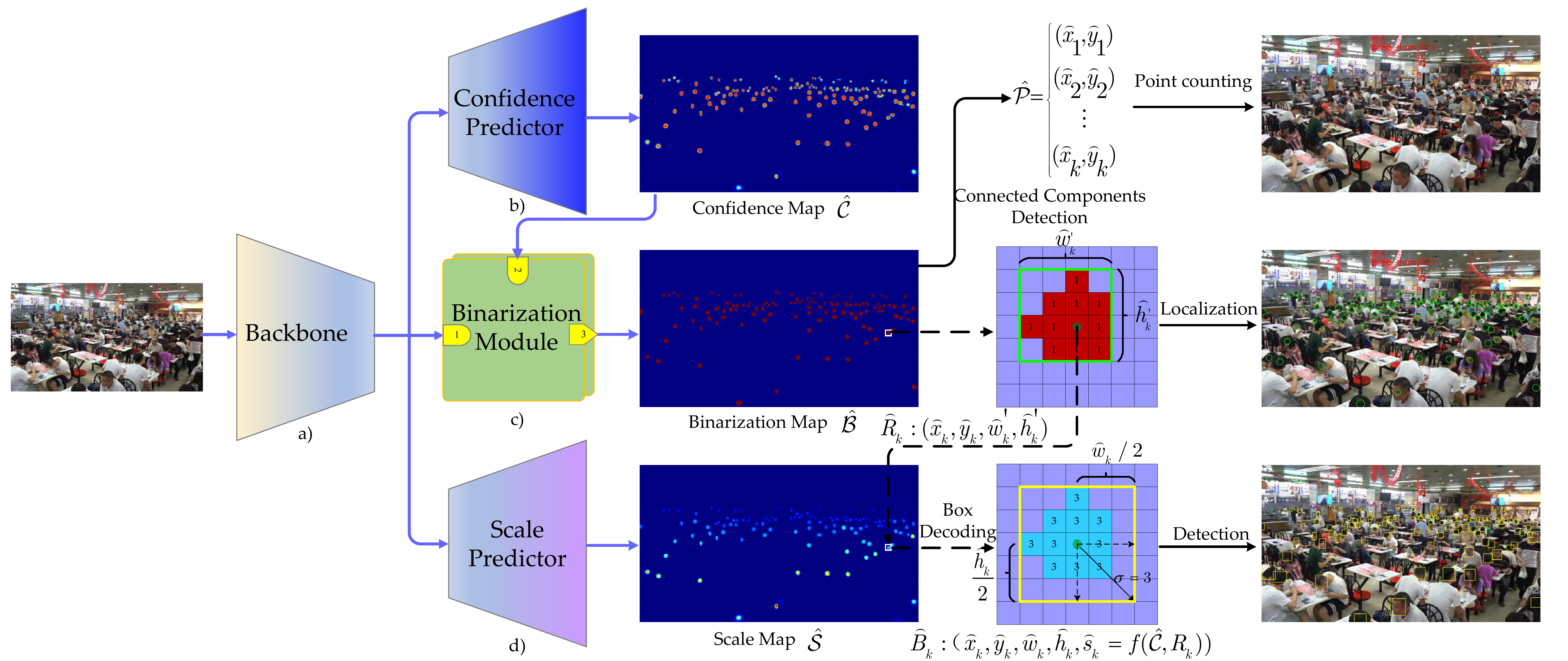}
	\caption{The skeleton of the LDC-Net. It first predicts the confidence of the crowd area. Then, a binarization module is applied to the confidence map for obtaining the binary map. The connected components in the binary map are detected to output the initial boxes and the head centers. Meanwhile, a scale predictor branch regresses the scale map to decode the accurate head size.}
\label{fig:framework}
\end{figure*}

\textbf{Point-supervision Crowd Localization.\,\,\,\,}
Besides the heat maps, some methods also directly exploit point labels to supervise the model for crowd localization. Laradji \emph{et al.} \cite{laradji2018blobs} combine four types of loss functions to push the model segment a single blob for each object. Liu \emph{et al.} \cite{liu2019point} propose a curriculum learning strategy to generate pseudo box-level labels from point annotations. Sam \emph{et al.} \cite{sam2020going} train a pixel-wise binary classifier to detect people instead of regressing local crowd density, which proposes a novel multi-scale architecture incorporating top-down feedback to address the scale variation. Furthermore, they propose the upgraded version, LSC-CNN \cite{sam2020locate}, which is a tailor-made dense object detection method that predicts position, size for each head simultaneously only with point-level annotations. Wang \emph{et al.} \cite{wang2020self} propose a self-training strategy to estimate the centers and sizes of crowded objects, which initializes pseudo object sizes from the point-level labels. However, these labels do not reflect head sizes well.

\subsection{Crowd Detection}
\label{sec:crowd_detection}
Crowd detection refers to detect people's heads in a crowd scene, which is a multi-view detection comparing with single-view face detection \cite{yang2016wider,najibi2017ssh,hu2017finding}. The head detection methods also can locate the position with a bounding box.

\textbf{Multi-view in Sparse Scenes.\,\,\,\,} In the last decade, Ishii \emph{et al.} \cite{ishii2004face} propose a head detector using hand-crafted features (four directional features, FDP) for a real-time surveillance system. Due to the limitation of hand-crafted features, some methods utilize CNN to improve complicated scenes' performance. Vu \emph{et al.} \cite{vu2015context} present a context-aware head detection algorithm based on R-CNN detector \cite{girshick2014rich} in natural scenes. Stewart and Andriluka \cite{stewart2016end} design a recurrent LSTM layer for sequence generation and an end-to-end detector (OverFeat \cite{sermanet2013overfeat}) to output a set of distinct detection results. However, these detectors work well in sparse scenes but fail in dense crowds.

\textbf{Single-view in Moderate Scenes.\,\,\,\,} As the WIDER FACE dataset \cite{yang2016wider} is proposed for face detection, some researchers design specific networks for relatively tiny objects. Hu and Ramanan \cite{hu2017finding} propose a tiny object detection framework by analyzing the impacts of image resolution, head scale, and context, significantly improving the performance in dense crowds. After this, there are many works \cite{bai2018finding,2019PyramidBox,deng2019retinaface,Li2020DSFD} focusing on this field. Among them, RetinaFace \cite{deng2019retinaface} extra annotates five facial landmarks on the WIDER FACE dataset and significantly improves the hard face detection. However, they are both designed for the single-view and lose many tiny instances in the extremely congested crowds due to the difficulty in extracting the facial feature of dense objects.

\textbf{Multi-view in Dense Scenes.\,\,\,\,} In recent years, some works try to detect head at multi-view on the point annotation crowd datasets. Liu \emph{et al.} \cite{liu2019point} simultaneously detect the size
and location heads in crowds, which initializes the pseudo ground truth bounding boxes from point-level annotations and then proposes an online updating scheme to refine the pseudo ground truth. LSC-CNN \cite{sam2020locate} adopts a per-pixel classification paradigm to size head, where a set of bounding boxes are fixed with predefined sizes, and then the model simply classifies each head to one of the boxes or as background. Crowd-SDNet \cite{wang2020self} initializes pseudo object sizes based on a locally uniform distribution assumption and proposes the confidence and order-aware refinement scheme to continuously refine the initial pseudo object sizes. FIDTM \cite{liang2021focal} proposes a focal inverse distance transform map to locate the head position and uses a KNN strategy to generate bounding boxes. Unfortunately, these methods can not size the head accurately for lacking the box level annotation.

\section{Proposed LDC-Net}
\label{sec:Proposed LDC-Net}

This work aims at addressing the above thorny matters that exist in the detection-based counting methods. As illustrated in Fig. \ref{fig:framework}, we establish a brand-new framework that can simultaneously tackle the detection, localization, and counting tasks in dense crowds. To make it easy to understand the proposed LDC-Net, Section \ref{Sec:Localization and Counting} and \ref{sec:detection} detail the methodology in addressing the three tasks. Section \ref{sec:loss_functions} introduces the criteria to supervise the proposed framework.

\subsection{Localization and Counting}
\label{Sec:Localization and Counting}
\subsubsection{Feature Extractor}
As shown in Fig. \ref{fig:framework} a), the feature extractor is exploited to generate the feature map from the input RGB image $\mathcal{I} \in \mathbb{R}^{H \times W \times 3}$, which can be any of the popular architectures. In this paper, we adopt two different backbones, VGG-16 \cite{simonyan2014very} and HRNet \cite{wang2020deep}. The former is commonly used in the crowd counting field, and the latter is a high-resolution network that can maintain the high-resolution representation as the network going deeply, which meets the LDC-Net's requirements well. For VGG-16, Feature Pyramid Network \cite{lin2017feature} is exploited to encode multi-scale features for improving the high-resolution representation ability, namely VGG-16+FPN. The feature extractor is defined as $\Phi$ with parameters $\theta_{e}$. Then the mapping can be represented as follow:


\begin{equation}
\mathcal{F} = \Phi (\mathcal{I}; \theta_{e}),
\label{Eq:1}
\end{equation}
where $\mathcal{F} \in \mathbb{R}^{c \times \frac{H}{4} \times \frac{W}{4}}$ ($c$ is the number of channels, which is 720 in HRnet and 768 in VGG-16+FPN. $H$ and $W$ are the width and height of the image.) is the high-level feature representation, which will be feed into confidence predictor, threshold learner and scale map predictor for subsequent tasks.

\subsubsection{Confidence Predictor}

Confidence predictor is a crucial module in LDC, which outputs the heat map for localization and counting. As shown in Fig. \ref{fig:framework} b), the purpose of  the confidence predictor is to decode the feature map $\mathcal{F}$ into a confidence map $\mathcal{\hat{C}} \in \mathbb{R}^{H \times W}$. From the visual intuition, the confidence map $\mathcal{\hat{C}}$ renders the object centers to individual blocks, and it will be applied to extract the object's localization.
\begin{figure}[htbp]
	\centering
	\includegraphics[width=0.45\textwidth]{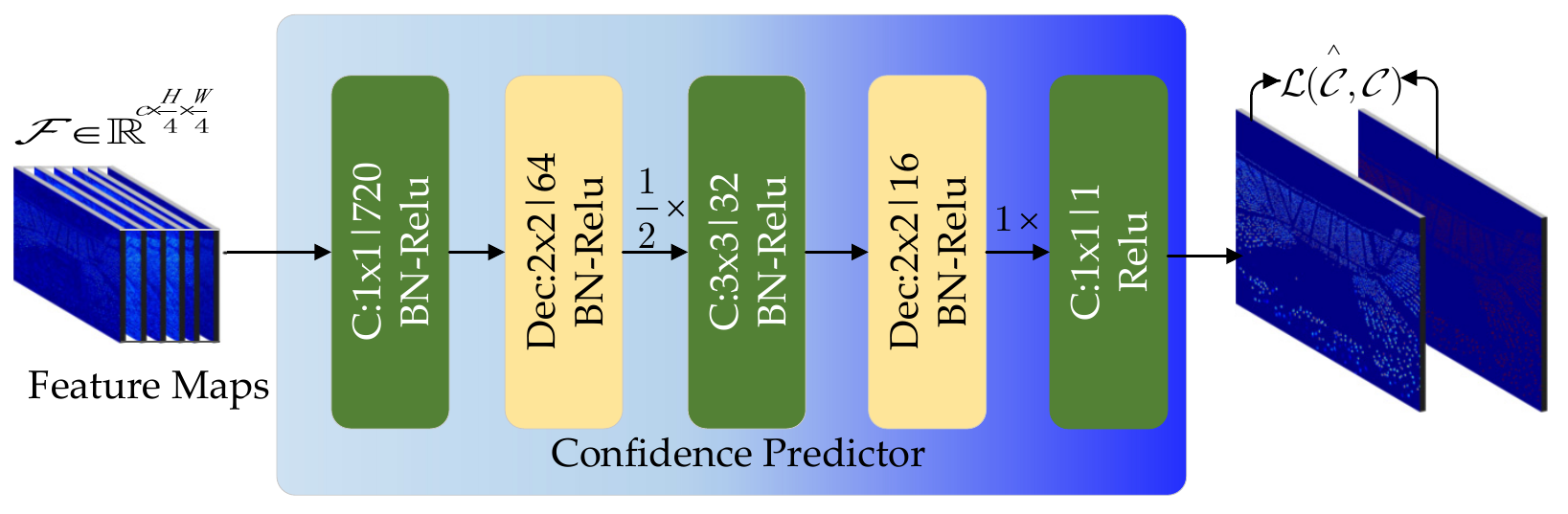}
	\caption{The composition of the confidence predictor. }
	\label{fig:confidence_predictor}
\end{figure}

As the confidence map's quality highly determines the localization performance, a fine-grained confidence map is urgently required to locate the head position accurately in \emph{dense}, \emph{scale diversity} and \emph{occlusion} scenes. To eliminate those adverse effects, we design a high-resolution representation decoder, whose configurations are illustrated in Fig.  \ref{fig:confidence_predictor}. For different backbones, the multi-scale representation $\mathcal{F}$ is first reintegrated to 720 channels by a $1 \times 1$ convolution layer. Then, a $2 \times 2$  transposed convolution layer with 2 stride increases the feature map size to $(\frac{H}{2},\frac{W}{2} )$, where the kernel is set to 2 for avoiding the checkerboard artifacts during upsampling. Moreover, a small convolution kernel that is friendly to the small targets can reduce the overlap of adjacent objects. Next, a $3 \times 3$ convolution layer reduces the feature channels to 32 for decreasing the computation. Then, another transposed convolution layer upsamples the size of feature maps to $(H,W)$, which is the same as the input image ${\mathcal{I}}$. Finally, a $1 \times 1$ convolution layer outputs the single-channel prediction as a confidence map. It is noticeable that each convolution layer and de-convolution layer are followed by a Batch Normalization layer and Relu activation function (the last layer is only followed by  Relu), which prevents overfitting and accelerates convergence speed. The confidence predictor is defined as $\Psi$ with parameters $\theta_{c}$, and the whole procedure is formulated with the following mapping:

\begin{equation}
\mathcal{\hat{C}} = \Psi (\mathcal{F}; \theta_{c}),
\label{Eq:2}
\end{equation}
where $\mathcal{\hat{C}}_{i,j} \geq 0  (1 \leq i \leq H, 1 \leq j \leq  W)$ represents the confidence value of position $(i,j)$ in a head region.

\subsubsection{Binarization Module}

\begin{figure*}[t]
	\centering
	\includegraphics[width=0.9\textwidth]{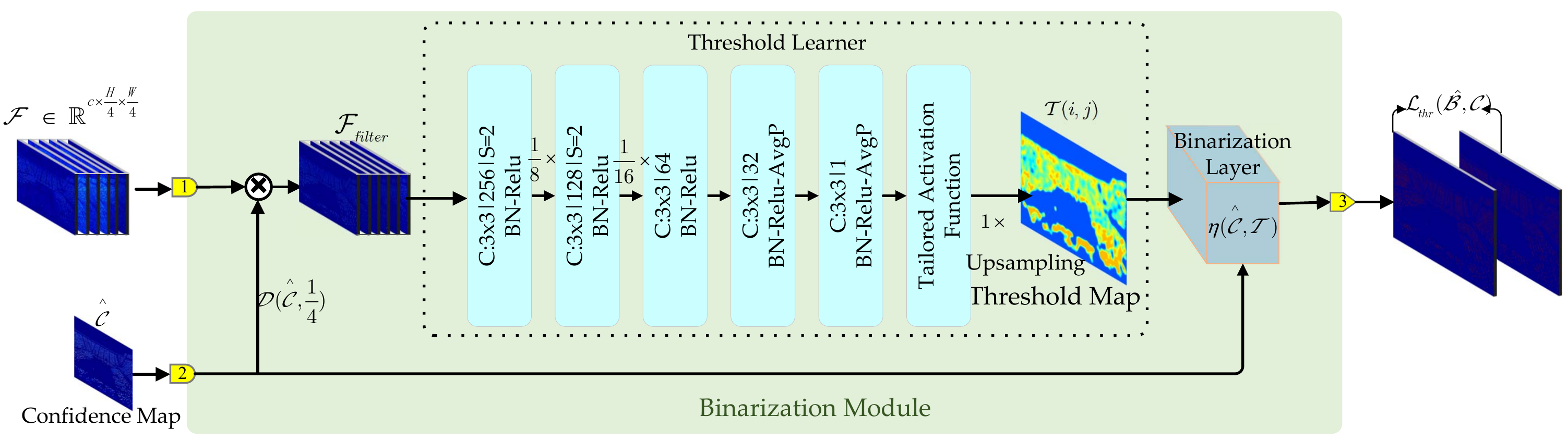}
	\caption{The flowchart of the pixel-level threshold learning process. It is designed to give proper thresholds for different confidence regions, making an optimal segmentation.}
	\label{fig:binarization_module}
\end{figure*}

 As mentioned in Sec. \ref{sec2:crowd_localizaiton}, many mainstream methods exploit heat maps for object localization. In general, a threshold is set to filter positional information from the predicted heat maps. Most heuristic crowd localization methods \cite{sam2020locate,abousamra2020localization,liang2021focal,wang2021self} extract the head points with a single threshold over a dataset. Transparently, it is not the optimal choice as the confidence responses are different between the hard (lower confidence) and relatively simple regions (higher confidence) as the confidence map shown in Fig. \ref{fig:framework} c). To alleviate this problem, IIM \cite{gao2020learning} proposes to learn a pixel-level threshold map to segment confidence map, which brings about effective ascension in catching more lower response heads and eliminating the overlap in the neighbor heads. But it also exists two problems: 1) The Threshold Learner \cite{gao2020learning} may induce a NaN (Not a Number) phenomenon during the training. 2) The predicted threshold map is relatively coarse. Here, we follow this concept and redesign the binarization module to tackle the two problems.

 As depicted in Fig. \ref{fig:binarization_module}, the feature map $\mathcal{F}$ plays another role in LDC-Net except for confidence prediction, which also is fed into a threshold learner for decoding a pixel-level threshold map. Here, we make a pixel-wise attention filter operation instead of directly conveying $\mathcal{F}$. The attention filter is a dot product operation between the original $\mathcal{F}$  and predicted confidence map $\hat{\mathcal{C}}$, which could be formulated as:

\begin{equation}
\mathcal{F}_{filter} = \mathcal{F} \otimes D (\mathcal{\hat{C}}, \frac{1}{4}),
\label{Eq:3}
\end{equation}
 where $D(x, y)$, a downsampling function, reduces the size of $x$ to $y \times$ of the input image. By this  production, $\mathcal{F}_{filter} \in \mathbb{R}^{c \times \frac{H}{4} \times \frac{W}{4}}$ is a an attention-guided feature representation, which is more efficient compared with directly forwarding feature $\mathcal{F}$.

The core components of the Binarization Module are the threshold learner and binarization layer. The former learns a pixel-level threshold map $\mathcal{T}$ from  $\mathcal{F}_{filter}$ and the latter binarizes the confidence map $\mathcal{\hat{C}}$ into a binary map $\mathcal{\hat{B}}$. As illustrated in Fig. \ref{fig:binarization_module}, the threshold learner is consists of five convolution layers: the first three layers progressively decrease the feature channels with $3 \times 3$ kernel size, and each of them followed by a Batch-Normalization and Relu activation function. The last two layers are with $3 \times 3$ and $1 \times 1$ kernel size, which are followed by Batch-Normalization, Relu, and Average Pooling layers. The Average Pooling layers with a window size of $9 \times 9$ are added to smooth the threshold map. Finally, a tailored activation function is introduced to resolve the NaN phenomenon arisen in IIM \cite{gao2020learning}, which is defined as:

\begin{equation}\operatorname{}f(x)=\left\{\begin{array}{ll}
0.25 & \text { if } x \leq 0.25 \\
0.90 & \text { if } x \geq 0.90 \\
x & \text { otherwise }
\end{array}\right. .
\label{Eq:4}
\end{equation}

Eq. \ref{Eq:4} limits the range of the $\mathcal{T}_{i,j}$ to $[0.25, 0.90]$. Compared with the Compressed Sigmoid \cite{gao2020learning} activation function, it does not force the last layer to output nonsense value (e.g., $\pm \infty$ in \cite{gao2020learning}) so it increases the stability of numerical. To assure the threshold is optimized properly in the training process, we stipulate the derivation rule as Eq. \ref{Eq:5},

\begin{equation}
\operatorname{}\frac{\partial f}{\partial x}=\left\{\begin{array}{ll}
e^{x-0.25} & \text { if } x<0.25 \\
0 & \text { if } x>0.95 \\
1 & \text { otherwise }
\end{array}\right. .
\label{Eq:5}
\end{equation}

The threshold learner is defined as $\phi$ with parameters $\theta_{t}$, which outputs the threshold map as the Eq. \ref{Eq:6} mapped,
\begin{equation}
\mathcal{T} = \phi (\mathcal{F}_{filter}; \theta_{t}),
\label{Eq:6}
\end{equation}
where $\mathcal{T}_{i,j} \in [0.25, 0.90], (1 \leq i \leq H, 1 \leq j \leq  W)$. Now, by forwarding the confidence map $\mathcal{\hat{C}}$ and threshold map $\mathcal{T}$ to the differentiable Binarization Layer \cite{gao2020learning}, we get the binary map $\mathcal{\hat{B}}$ with the function $\eta(\mathcal{\hat{C}}, \mathcal{T})$, which is formulated as:

\begin{equation}
\mathcal{\hat{B}}_{i,j}=\operatorname{\eta}(\mathcal{\hat{C}},\mathcal{T})=\left\{\begin{array}{ll}
1, & \text { if } \mathcal{\hat{C}}_{i,j} \geq \mathcal{T}_{i,j} \\
0, & \text { otherwise }
\end{array}. \right.
\label{Eq:7}
\end{equation}

The purpose of the binarization module is to optimize the threshold learner $\phi(\cdot)$, which outputs an adaptive threshold map for accurately binarizing the confidence map. However, there is no label to directly supervise the learning of $\mathcal{T}$. The solution is to use the mask map generated with the box/point annotation to supervise the binarization layer's output $\mathcal{\hat{B}}$, which means the gradient must be back-forward through the binarization layer. Eq. \ref{Eq:7} shows a discrete operation, and we use the derivative method proposed in IIM \cite{gao2020learning}.

\subsubsection{Connected Components Detection}
\label{sec:connected components detection}
After getting the binary map $\mathcal{\hat{B}}$, the localization and counting are equivalent to detect the connected component from $\mathcal{\hat{B}}$, where each blob corresponds to an individual. Let $\hat{R} = \{\left(\hat{x}_{k}, \hat{y}_{k}, \hat{w}_{k}^{\prime}, \hat{h}_{k}^{\prime}\right) |(k= 1 \cdots K)\}$ is the connected components set that contains $K$ blobs detected from the binary map, where $(\hat{x}_{k}, \hat{y}_{k})$ is the object center point, $\hat{w}_{k}^{\prime}$ and $\hat{h}_{k}^{\prime}$ are the blob's width and height. Then,  the points set $\hat{\mathcal{P}} = \{\left(\hat{x}_{k}, \hat{y}_{k}\right) |(k= 1 \cdots K)\}$ is the \emph{localization} result, and the number of the points is regarded as the \emph{counting} result.

\subsection{Detection}
\label{sec:detection}
In this section, we propose an anchor-free head detection algorithm that can accurately detect tiny and congested heads. As illustrated in Fig. \ref{fig:framework} d), a scale prediction branch is designed to regress the head size of a crowd scene. With the location information extracted in Sec. \ref{Sec:Localization and Counting} and the prediction scale map, we solve a bounding box for each object.

\subsubsection{Scale Predictor}
\begin{figure}[t]
	\centering
	\includegraphics[width=0.45\textwidth]{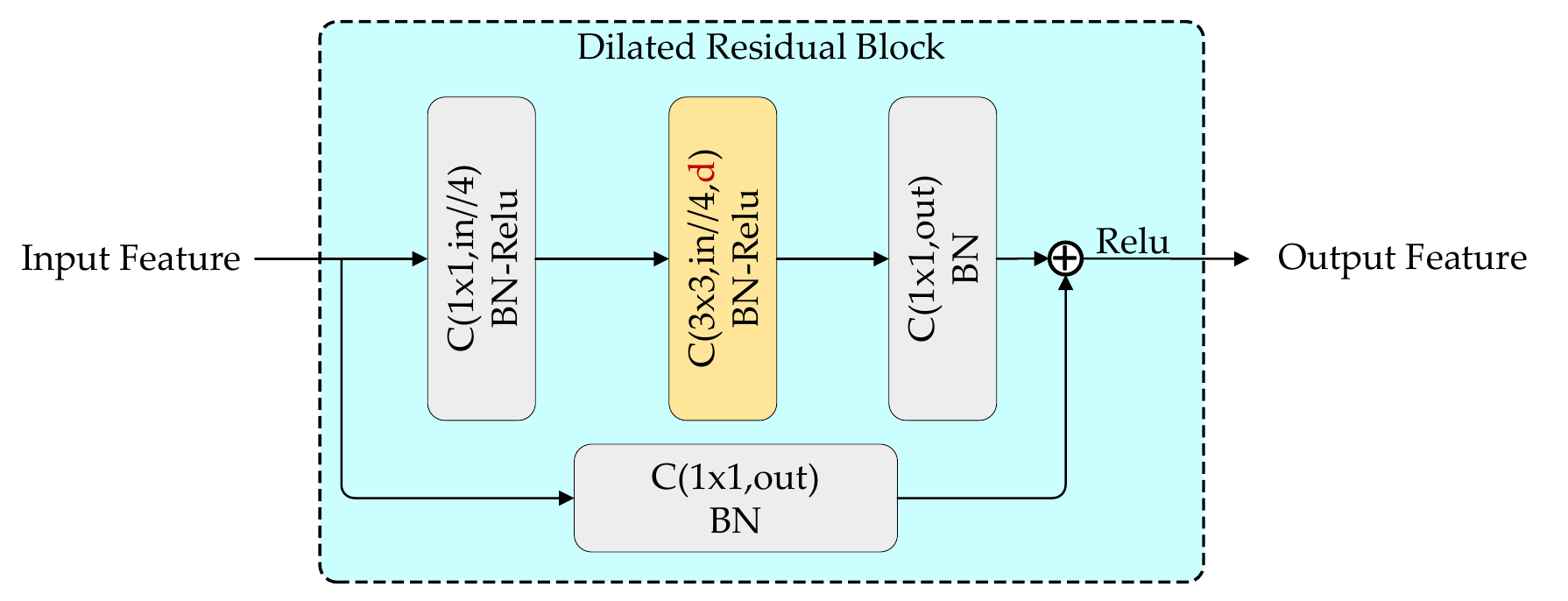}
	\caption{The architecture of dilated residual block.``in" and ``out" represent the input and output channels. respectively. ``d" is the dilated factor.}
	\label{fig:dilated_residual}
\end{figure}

 Scale predictor decodes the feature map $\mathcal{F}$ into a scale map, which has the same shape as the input image. To enhance the regression ability across a large-scale range, we propose a dilated residual block whose structure depicts in Fig. \ref{fig:dilated_residual}. Compared with the traditional residual block, the main contribution is to add the dilated operation in the middle $3 \times 3$  convolution layer. By setting different dilated ratios, we can increase the reception field to catch objects with different scales.

\begin{figure}[htbp]
	\centering
	\includegraphics[width=0.5\textwidth]{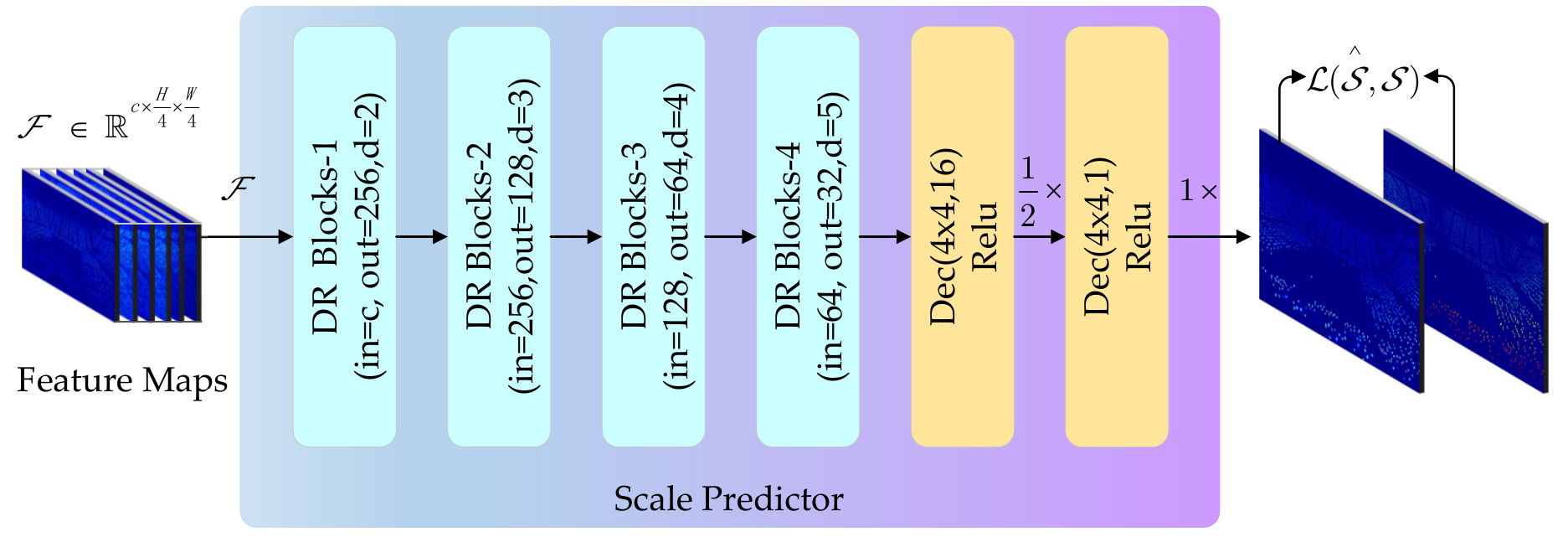}
	\caption{The skeleton of scale predictor. }
	\label{fig:scale_predictor}
\end{figure}

Fig. \ref{fig:scale_predictor} shows the specific components of the scale predictor. We first connect four dilated residual blocks consecutively by increasing the dilated rate from 2 to 5. This structure fuses multi-size reception fields and makes it possible to deal with the heads with a wide range of sizes.  Finally, two transposed convolution layers refine the feature to a one-channel scale map and recover the resolution to $H \times W$. The scale predictor is defined as $\psi$ with the parameter $\theta_{s}$, and the scale map is mapped as:

\begin{equation}
\mathcal{\hat{S}} = \psi (\mathcal{F}; \theta_{s}),
\label{Eq:8}
\end{equation}
where $\mathcal{\hat{S}} \in \mathbb{R}^{1\times H \times W}$ has the same shape with confidence map and binarization map. They decide the bounding box together.

As shown in Fig. \ref{fig:framework} and mentioned in Sec. \ref{sec:connected components detection}, for the $k$-th object, the center $\mathcal{\hat{P}}_{k} = (\hat{x}_{k}, \hat{y}_{k})$ and the initial size $(\hat{w}_{k}^{\prime}, \hat{h}_{k}^{\prime})$ is first detected from the binary map. With the initial size information, we decode the angle between the width and the diagonal, which is formulated by the trigonometric functions:

\begin{equation}
\left\{\begin{array}{ll}
 \sin\alpha = \hat{h}_{k}^{\prime}/ \sigma^{\prime}_{k}\\
 \cos\alpha  = \hat{w}_{k}^{\prime}/ \sigma^{\prime}_{k}
\end{array},\right.
\label{Eq:9}
\end{equation}
where $\hat{\sigma}_{k}^{\prime}=\sqrt{\hat{h}_{k}^{\prime 2}+\hat{w}_{k}^{\prime 2}}$ is the diagonal length and $\alpha$ is the angle. The more accurate size $\hat{\sigma}_{k}$ is given by querying the corresponding value of the center point $\hat{\mathcal{P}}_{k}$ in the scale map $\mathcal{\hat{S}}$, which is formulated as:

\begin{equation}
\hat{\sigma}_{k}={\mathcal{\hat{S}}(\hat{x}_{k},\hat{y}_{k})}.
\label{Eq:10}
\end{equation}
With center point and $\hat{\sigma}_{k}$, we can produce a bounding box for each head. To evaluate the detection performance with the universal metrics (e.g., AP), we exploit the confidence map  $\mathcal{C}$  to give each bounding box a score within its connect component region. Then the detection result is represented by the following tuple:\\
\centerline{$(\hat{x}_{k}-\hat{w}_{k}, \hat{y}_{k}-\hat{h}_{k}, \hat{x}_{k}+\hat{w}_{k}, \hat{y}_{k}+\hat{h}_{k},  \hat{s}_{k})$} \\
where $\hat{w}_{k} = \sigma_{k}\cos\alpha $ and $\hat{h}_{k} =  \sigma_{k}\sin\alpha$ are the final box size. $\hat{s}_{k}$ is given by Eq. \ref{Eq:11},

\begin{equation}
 \hat{s}_{k} = {\mathcal{\hat{C}}(\hat{x}_{k},\hat{y}_{k})}.
 \label{Eq:11}
\end{equation}

\subsection{Training for LDC-Net }
\label{sec:loss_functions}
\subsubsection{Ground Truth Generation}
As introduced in Sec. \ref{sec:introduction}, there are several types of heat maps proposed for localization.  In the object detection field, CenterNet \cite{zhou2019objects} is the first work that regresses the heat map for object centers. It treats the object center points as 1 and the other pixels as 0,  then splats the ground truth points into a heat map by a 2D size-adaptive Gaussian kernel. Similar to CenterNet, in the crowd understanding field, LSC-CNN \cite{sam2020locate} manually predefines some head size to generate multi-scale size heat maps at head centers for localization, which is a pioneering work that achieves head bounding box prediction without box annotations. However, it only gives the coarse box prediction because the heat map of size is not accurate with the supervision of the predefined box bins (12 sorts of size). Besides, Crowd-SDNet \cite{wang2021self} generates the center points heat map and pseudo size map for localization and box prediction. FIDTM \cite{liang2021focal} generates a heat map by performing the distance transform function on the dot map, which avoids overlap caused by the Gaussian blobs but treats all heads as the same scale. To avoid overlap and take count of the scale information to some extent, TopoCount \cite{abousamra2020localization} dilates the dot map up to 7 pixels by determining the distance of the nearest neighbor dot. IIM \cite{gao2020learning} proposes to generate the Independent Instance Map (IIM) guided by box size, which is more reasonable and effective. Here, we follow the IIM and refine its generation method further. We first introduce the point-guide IIM and then generate the box-guide IIM as the label by injecting the box information.

\textbf{Confidence Map $\mathcal{C}$ (Point-guide).}\quad Let $\mathcal{P} = \{(x_{i},y_{i})\}_{i=1}^{N}$ be the point-wise annotations in a crowd image,  where $N$ is the number of heads in the image. Following the TopoCount \cite{abousamra2020localization}, we use KNN algorithm to build the distance tree for the 2D coordinates. Then, For the $i$-th object $\mathcal{P}_{i}$ in the image, we query the nearest neighbor $\mathcal{P}_{j}$. The dilated size of the $i$-th object is defined as Eq. \ref{Eq:12},
\begin{equation}
w^{{p}}_{i} = h^{p}_{i} = \text{min}(c, r*\text{dist}(\mathcal{P}_{i}, \mathcal{P}_{j})),
\label{Eq:12}
\end{equation}
where $\text{dist}(\mathcal{P}_{i}, \mathcal{P}_{j}) = \sqrt{\left(x_{i}-x_{j}\right)^{2}+\left(y_{i}-y_{j}\right)^{2}}$ is the Euclidean Distance between the two points. $c$ is a constant value that limits the mask size of object. It is a little different for each dataset and we reports the setting in Table \ref{Table:dilated_size}. $r$ is a factor that controls two adjacent blobs not to overlap, and it is set to 0.25 in our experiments.

\begin{table}[htbp]	
\centering
\caption{The configuration of maximum dilated size in different datasets.}
\scriptsize
	\setlength{\tabcolsep}{1.1mm}{\begin{tabular}{cIc|c|c|c|c}
		\whline
		Dataset&Shanghai Tech & UCF-QNRF & NWPU-Crowd & FDST & WIDER FACE  \\
		\hline
		$c$&15& 30& 30& 30& 50 	\\
		\whline		
	\end{tabular}}\label{Table:dilated_size}
\end{table}

\textbf{Confidence Map $\mathcal{C}$ (Box-guide).}\quad Recently, some crowd datasets (e.g., NWPU-Crowd \cite{gao2020nwpu} and FDST\cite{fang2019locality}) provide the box annotation. It is a more reasonable and precise way to generate IIM by taking into account the box annotation. Let $\mathcal{B} = \{x_{lt}^{i},y_{lt}^{i}, x_{br}^{i},y_{br}^{i}\}_{i=1}^{N}$ be the bounding box annotations in a crowd image. For the $i$-th object in the image, the dilated size of the $i$-th object is determined by its box size $\mathcal{B}_{i}$ and the nearest neighbor $\mathcal{P}_{j}$, which is defined as Eq. \ref{Eq:13},

\begin{equation}
\small
(w^{b}_{i}, h^{b}_{i}) = \left\{\begin{array}{ll}
 (\text{min}(w_{i}^{p}, w_{i}), \text{min}(w_{i}^{p}, w_{i})/w_{i}h_{i} ) & \text{ if }  w_{i} > h_{i}\\
 ( \text{min}(h_{i}^{p}, h_{i})/h_{i}w_{i},  \text{min}(h_{i}^{p}, h_{i}) ) &  \quad \text{else}
\end{array},\right.
\label{Eq:13}
\end{equation}
where $w_{i} = x_{br}^{i} - x_{lt}^{i}$ and $ h_{i} = y_{br}^{i} - y_{lt}^{i} $ are the box width and height of $i$-th object. Eq. \ref{Eq:13} gives more accurate scale information for IIM with considering the bounding box size and ratio. Especially, the box-guide IIM generates more accurate blobs in the spare scenes, in which the head size is hard to describe by the KNN algorithm.

\textbf{Size Map $\mathcal{S}$}.\quad The size map is inherited from the confidence map, which is generated by replacing the head mask value with the size value for each individual blob. We define the size of the head by the following formulation,
\begin{equation}
\sigma_{i} = \sqrt{w_{i}^{2}+ h_{i}^{2}}/2.
\label{Eq:14}
\end{equation}
where $\sigma_{i}$ represents the scale information for $i$-th object, which is the distance from the center to box corner.
\subsubsection{Loss Function}
There are four criteria to supervise the LDC-Net. Among them, $\mathcal{L}_{con}$ and $\mathcal{L}_{OUSR}$ supervise the feature extractor and confidence predictor. $\mathcal{L}_{thr}$ guides the learning of the binarization module, and the rest $\mathcal{L}_{size}$ is set for scale predictor.

\emph{Confidence Regression Loss.\quad} The prediction confidence map $\mathcal{\hat{C}}$ is trained with an L2 loss:
\begin{equation}
\mathcal{L}_{con}(\mathcal{\hat{C}}, \mathcal{C}) = \frac{1}{2N} \sum_{k=1}^{N}\|\mathcal{C}_{k}-\hat{C}_{k}\|_{2}^{2},
\label{Eq:15}
\end{equation}
where $N$ is the batch size, $\hat{C}_{k}$ and  $C_{k}$  represent the prediction confidence map and ground truth map of $k$-th sample in a batch, respectively.

\emph{Threshold Regression Loss:} The threshold map $\mathcal{T}$ is trained by an L1 loss:

\begin{equation}
\mathcal{L}_{thr}(\mathcal{\hat{B}}, \mathcal{C}) = \frac{1}{N} \sum_{k=1}^{N}\|\mathcal{C}_{k}-\hat{B}_{k}\|,
\label{Eq:16}
\end{equation}
where $\mathcal{\hat{B}}_{k}$ generated with Eq. \ref{Eq:7} is the binarization resultant of the $i$th image. Eq. \ref{Eq:16} reveals that we use the same ground truth, confidence maps, to train threshold learner. It does not consume extra annotation labor so that this strategy can be applied in the existing datasets.

\emph{Over and Under Segmentation Ratio Loss.\quad} The contents of the preceding parts illuminate that the LDC-net can output an independent binarization map end-to-end, of which the quality of binarization will directly determine the localization and counting performance. According to the experimentation, the network tends to be partial to background as the head region has a small ratio under the supervision of L2 loss, which causes the small objects to be partially detected. To further catch the failure cases, we propose an Over and Under Segmentation Ratio loss (OUSR loss for short) to increase the foreground region's regression. The OUSR loss is directly conducted on two binary maps, predicted $\mathcal{\hat{B}}$ and the ground truth $\mathcal{C}$, which is formulated as:

\begin{equation}
\mathcal{L}_{OUSR}(\mathcal{\hat{B}}, \mathcal{C}) = \frac{\sum  \mathcal{\hat{B}} \circ (1-\mathcal{C})}{\sum   (1-\mathcal{C})+\varepsilon} + \frac{ \sum |\mathcal{C}-\mathcal{\hat{B}}| \circ \mathcal{C}}{\sum  \mathcal{C}+\varepsilon} ),
\label{Eq:17}
\end{equation}
where $\circ$ is the Hadamard product. The first term in Eq. \ref{Eq:17} describes the background segmentation rate, and the second term is the foreground segmentation rate. $\varepsilon$ is a very small value to avoid nonsense division (e.g., a ground truth $\mathcal{C}$ without object). Ideally, both of them are expected to be 0 in a batch. By using this function, the foreground and background pixels are adaptively given different weights. Especially in the sparse scenes, the foreground pixels have larger weights, while the background pixels have smaller weights.

\emph{Size Regression Loss.} This loss is to supervise the scale predictor by performing L1 function between the predict size map $\mathcal{\hat{S}}$ and the ground truth map $S$,
\begin{equation}
\mathcal{L}_{size}(\mathcal{\hat{S}}, \mathcal{S}) = \frac{1}{N} \sum_{k=1}^{N}\|\mathcal{S}_{k}-\hat{S}_{k}\|.
\label{Eq:11}
\end{equation}

Finally, the whole LDC-Net is trained with the weighted loss:

\begin{equation}
\mathcal{L} = \mathcal{L}_{con} +  \mathcal{L}_{thr} + \lambda \mathcal{L}_{OUSR} + \mathcal{L}_{size}.
\label{Eq:19}
\end{equation}
The $\lambda$ is set to 0.01 in experiments except for the specified otherwise. We analyze its sensitivity in Sec. \ref{sec:ousr_loss}.

\section{Experiments}
\label{sec:Experiments}
\subsection{Dataset}
Five mainstream datasets are exploited to evaluate the LDC-Net. Noticeably, the localization and counting performance evaluation is conducted over all datasets as it only requires point annotation. The detection performance is demonstrated on the NWPU-Crowd \cite{gao2020nwpu}, FDST \cite{fang2019locality}, and WIDER FACE \cite{yang2016wider} datasets as they provide the box-level annotation.

\textbf{NWPU-Crowd} \cite{gao2020nwpu} benchmark is established in 2020. It is the largest and most challenging open-source crowd dataset at present. It contains head points and box labels. There are a total of 5,109 images and 2,133,238 annotated instances.

\textbf{Shanghai Tech} \cite{zhang2016single} has two subsets: \textbf{Part A} has 482 images (300 images for training, 182 images for testing), with a total of 241,677 instances, and \textbf{Part B} contains 716 images (400 images for training, 316 images for testing), including 88,488 labeled heads.

\textbf{UCF-QNRF} \cite{idrees2018composition} is a large-scale dense crowd dataset, consisting of 1,535 images (1,201 images for training and 334 images for testing), with a total of 1,251,642 instances.

\textbf{FDST} \cite{fang2019locality} consists of 100 video sequences that are collected from 13 different scenes, which contains 15,000 frames (9,000 images/60 sequences for training and 6,000image/40 sequences for testing), with a total of 394,081 heads, including the point and box labels.

\textbf{WIDER FACE} \cite{yang2016wider} is a face detection dataset, which collects 32,203 images and annotates 393,703 faces with bounding boxes. WIDER FACE dataset is organized based on 61 event classes. Each event class is selected $40\%/10\%/50\%$ data as training, validation, and testing sets.

\subsection{Implementation Details}
 In data augment, we use the random horizontally flipping, scaling ($0.8 \times\sim1.20\times$), and cropping (768$\times$1024) strategies. The batch size is 4 in Shanghai Tech and 8 in other datasets. The learning rates of learnable modules are initialized to $5e-5$ except for the Binarization Module, which is $1e-5$. During training, the learning rate is updated by a decay strategy, $new\_lr = base\_lr \times {(1-current\_iteraiton/all\_iteration)}^{0.9}$. Adam \cite{kingma2014adam} algorithm is adopted to optimize the framework. The backbones, HR-Net and VGG-16, are initialized with the pre-trained weights on ImageNet \cite{krizhevsky2012imagenet}. We divide the 10\% training dataset as a validation set except for the NWPU-Crowd, FDST, and WIDER FACE as they have provided the validation set.  In the test phase, we select the best-performing model on the validation set to evaluate the performance on the testing set. We perform an end-to-end prediction without any multi-scale prediction fusion and parameters search. 

\subsection{Metrics}
To make a comprehensive and fair comparison with the previous work, we review the different metrics and evaluate them uniformly in this paper.

\subsubsection{Localization Metrics}

\textbf{AP/AR/F1-score} are introduced by Idrees \emph{et al.} \cite{idrees2018composition} and Abousamra \emph{et al.} \cite{abousamra2020localization}, which evaluate the average precision, average recall, and F-score of localization results at multiple distance thresholds (1,2,3,$\cdots$,100 pixels). The estimated points are associated with the ground truth by a one-to-one matching mechanism, and then matched pairs within the distance threshold are treated as Truth Positives (TP). Otherwise, they are False Positives(FP). If the prediction points have no associated points, they are be regarded as False Negatives (FN).

\textbf{MLE} \cite{sam2020locate} is proposed to measure the Mean Localization Error between the annotated and predicted points, which first match the prediction points to the annotation points in a one-to-one pattern, and then sum the distance of matched pair. Besides, there would be some redundant or deficient prediction points for an image. They are given a fixed distance penalty of 16 pixels. Finally, the MLE is an average value over all target points in a testing set.

\textbf{Pre/Rec/F1-m} are utilized to evaluate the performance in crowd localization challenge, NWPU-Crowd \cite{gao2020nwpu}, which evaluates the model by calculating the Precision (Pre.), Recall (Rec.), and F1-measure (F1-m) over the whole instances in a dataset. The prediction points are assigned to the target points with one-to-one matching. Truth positive points are these fall into the bounding box determined cycle regions. The radius $\sigma_l$ is calculated with box width and height: $\sigma_l={\sqrt {{w^2} + {h^2}} }/2 $, where $w$ and $h$ are the width and height of the instance, respectively.

It is noticeable that the first and second metrics can apply to both point-annotated datasets and box-annotated datasets. The third metric requires the box label. For the datasets that only have point annotation, we use the pseudo box generated by IIM \cite{gao2020learning}.

\subsubsection{Counting Metrics}
For the counting task, the distance between the predicted counts and the ground-truth ones are measured by the widely-used metrics, Mean Absolute Error (MAE), Root Mean Square Error (RMSE), and mean Normalized Absolute Error (NAE),
\begin{equation}
\tiny
M A E=\frac{1}{N} \sum_{i=1}^{N}\left|y_{i}-\hat{y}_{i}\right|, M S E=\sqrt{\frac{1}{N} \sum_{i=1}^{N}\left|y_{i}-\hat{y}_{i}\right|^{2}}, N A E=\frac{1}{N} \sum_{i=1}^{N} \frac{\left|y_{i}-\hat{y}_{i}\right|}{y_{i}}.
\end{equation}
where $N$ is the number of images, $y_i$ is the counting value and $\hat{y}_i$ is the estimated value for the $i$-th test image.

\subsubsection{Detection Metrics}
  The detection task follows the WIDER FACE \cite{yang2016wider} evaluation protocol, which uses Average Precision (AP) to measure the detection performance, in which the true positives are those intersections of union (IOU) between the prediction and target boxes greater than a threshold (e.g., IOU$\geq$0.5).
\begin{table*}[htbp]	
	\centering
	\caption{Ablation study for the OUSR loss on Shanghai Tech Part B dataset. We divide 10$\%$ training data as  \emph{val set}. $\lambda$ is the weight of OUSR loss. ``*'': The authors do not provide the results, and we calculate them with its pre-trained model. }
	\setlength{\tabcolsep}{0.95mm}{\begin{tabular}{cIcIcIc|c|cIc|c|cIcIc|c|cIc|c|c}
			\whline
			\multirow{3}{*}{Method}  & \multirow{3}{*}{$\lambda$} &\multicolumn{7}{cI}{Val set} &\multicolumn{7}{c}{Test set} \\
           \cline{3-16}
			  &  &\multicolumn{4}{cI}{Localization $\uparrow$ } &\multicolumn{3}{cI}{Counting $\downarrow$ } &\multicolumn{4}{cI}{Localization $\uparrow$ } &\multicolumn{3}{c}{Counting $\downarrow$ }\\
			\cline{3-16}
			&     &\textbf{MLE}&\textbf{F1-score} & AP & AR &MAE &MSE&NAE &\textbf{MLE}&\textbf{F1-score} & AP & AR &MAE &MSE&NAE	\\
			\whline
			\multirow{1}{*}{LSC-CNN \cite{sam2020locate}} &- &- &- & - &-&- &- &-	&9.0&86.7*  &87.3* &86.1*& 8.1 &\textbf{12.7}&0.069*\\
			\whline
		    \multirow{3}{*}{LDC-Net}
                  &0     &7.8   &90.3  &90.3&90.3         &8.0 &13.8 &0.068   &8.1 &90.2 &89.5&91.0    &8.9 &14.4 &0.074	\\
			\cline{2-16}
			       &0.1  &\textbf{7.5}   &90.2  &\textbf{90.6} &90.4        &8.3&13.6 &0.065    &7.9 &90.0 &89.4&90.7   &9.4 &16.2 &0.075	\\
			\cline{2-16}
			       &0.01 &7.6   &\textbf{90.7} &90.5&\textbf{91.0}          &\textbf{7.5} &\textbf{12.7} &\textbf{0.063}   &\textbf{7.7} &\textbf{90.4}&\textbf{90.8}&\textbf{91.3}   &\textbf{8.1} &13.5 & \textbf{0.066}	\\

			\whline			
	\end{tabular}}
	\label{Table:OUSR_loss}
\end{table*}

\subsection{Ablation Study}
This section is conducted to discuss the effectiveness of the binarization module and the proposed OUSR loss. For the binarization module, we compare it with the fixed thresholds. For OUSR loss, we explore how it improves performance. Besides,  a group of comparison experiments is conducted to choose a proper hyper-parameter for it.

\subsubsection{Effectiveness of Learnable Thresholds}

\begin{table}[tbp]	
	\centering
	\caption{Ablation study on different binarization methods. The reported results are tested on the NWPU-Crowd \emph{val set} since the test set labels are not available.}
	\setlength{\tabcolsep}{0.95mm}{\begin{tabular}{c|cIc|c|cIc|c|c}
			\whline
			\multirow{2}{*}{Method}  & Threshold &\multicolumn{3}{cI}{Localization $\uparrow$ } &\multicolumn{3}{c}{Counting $\downarrow$ } \\
			\cline{3-8}
			&type     &\textbf{F1-m} & Pre & Rec &MAE &MSE&NAE	\\
			\whline
			\multirow{3}{*}{IIM \cite{gao2020learning}} &fixed:0.5  &74.2 &92.4 & 62.0 &131.4&645.5 &0.237	\\
			\cline{2-8}
			  &fixed:0.8    &62.6   &94.1   &46.9 &197.4 &734.8 &0.420 	\\
			\cline{2-8}
			 &learnable   &80.2 &84.1 &76.6 &55.6&330.9 &0.197	\\
			\whline
		    \multirow{3}{*}{LDC-Net}
                  &fixed:0.25  &78.8 &81.7 &76.1&52.7&215.4 &0.155	\\
			\cline{2-8}
			       &fixed:0.3  &80.4 &83.3 & 77.7	&47.4&161.8 &\textbf{0.147}\\
			\cline{2-8}
			       &fixed:0.35  &80.9 &\textbf{84.9}  &77.3 &50.8 &191.9 &0.149\\
            \cline{2-8}
			       &learnable  &\textbf{81.3} &81.5  &\textbf{81.2} &\textbf{39.2} &\textbf{93.7} &0.150\\

			\whline			
	\end{tabular}}
	\label{Table:ablation}
\end{table}

Some previous methods \cite{sam2020locate, wang2021self} perform the parameter search to find the best threshold by minimizing the counting error on the \emph{val set}. Here, we use this strategy to find the optimal fixed threshold and compare its performance with the learnable thresholds. The experiments are conducted on the NWPU-Crowd benchmark and evaluated on the localization and counting tasks. Table \ref{Table:ablation} shows that the methods equipped with learnable thresholds have a significant improvement in localization and counting performance compared with the fixed threshold. For the fixed threshold strategy, with improving the threshold value, the localization precision increases while other metrics decline. In the learnable threshold scheme, all metrics are trade-off as the threshold is adaptive for each instance. As shown in the last row, the learnable thresholds achieve state-of-the-art in both counting and localization. Especially in the counting task, the MAE and MAE  improve  $17.3\%$ and $42.1\%$, respectively.

\begin{figure}[htbp]
	\centering
	\includegraphics[width=0.5\textwidth]{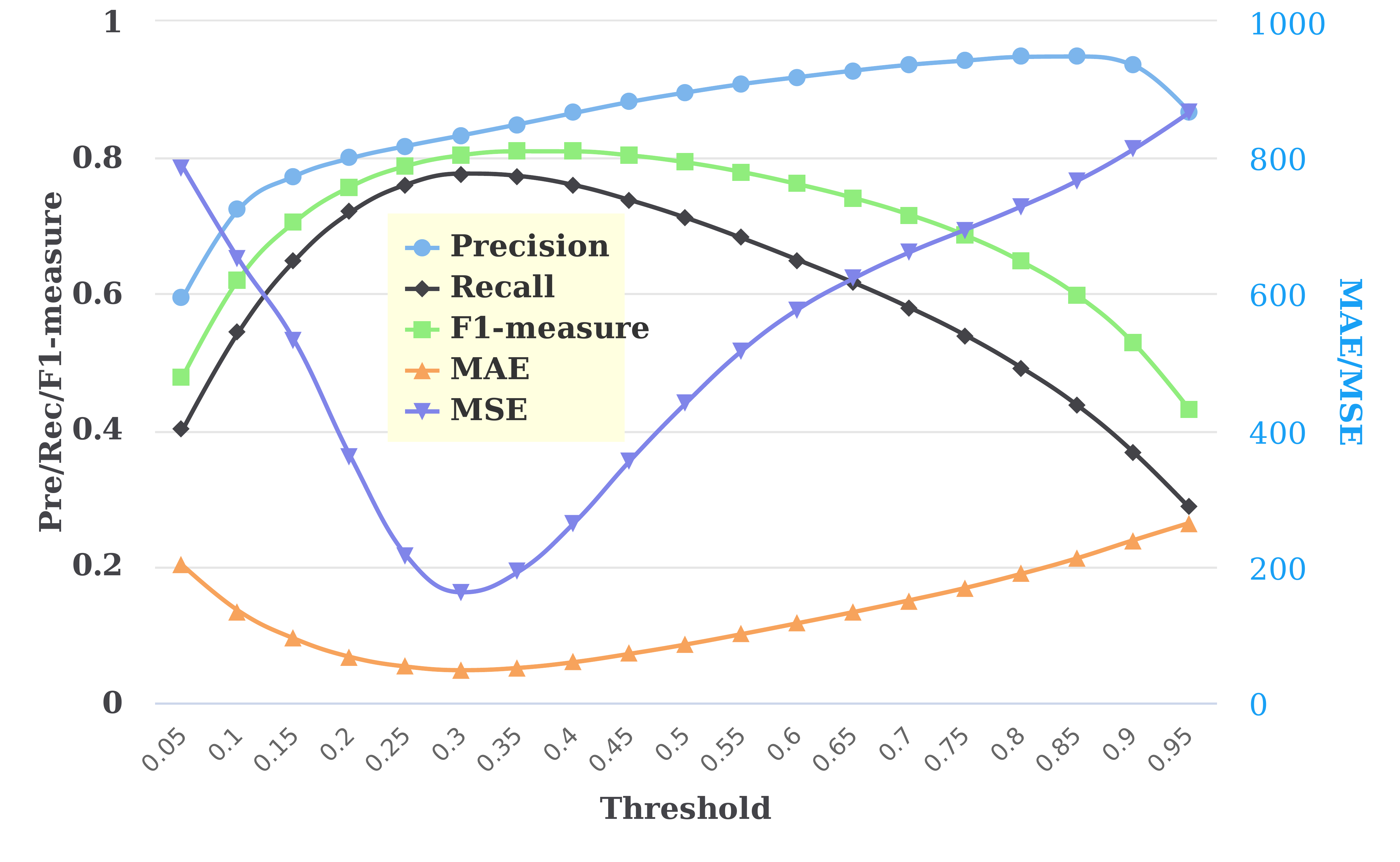}
	\caption{The localization and counting performance on the NWPU-Crowd \emph{val set}. These results are calculated by replacing the learnable binarization module with different fixed thresholds in a best-performance model.}	\label{fig:fixed_thresholds}
\end{figure}

Besides, we also draw the metric curves to show the trend in Fig. \ref{fig:fixed_thresholds}. As increasing the threshold from 0.05 to 0.30 with a step of 0.05, all metrics have a rapid upward trend. The best counting (MAE/MSE:\quad 47.4/161.8) and localization (F1-m/Pre/Rec:\quad 80.9/84.9/77.3) results appear when the threshold around 0.30 and 0.35, respectively. From 0.40 to 0.95, the metrics start to go down. Although we can perform such a parameter search to find the optimal threshold for a trained model, it not the optimal threshold over the whole process. The binarization module provides an automatic way to substitute cumbersome searching. And it can set thresholds according to the image content to gets a better result.

\begin{table*}[htbp]
	\centering
	\small
	\caption{The leaderboard of NWPU-Crowd Localization (\emph{test set}). $A0 \sim A5$ respectively means that the head area is in  $[10^0,10^1]$, $(10^1,10^2]$,  $(10^2,10^3]$,  $(10^3,10^4]$,  $(10^4,10^5]$, and $>10^5$. ``*'' represents anonymous submission on the benchmark. Some results are recorded in \url{https://www.crowdbenchmark.com/historyresultl.html}.}
	\begin{tabular}{cIcIc|cIc|c}
		\whline
		\multirow{2}{*}{Method}	&\multirow{2}{*}{Backbone}  &\multicolumn{2}{cI}{Overall ($\sigma_l$)}  &\multicolumn{2}{c}{Box Level (only Rec under $\sigma_l$) (\%)}  \\
		\cline{3-6}
		&  & \textbf{F1-m}/Pre/Rec (\%) $\uparrow$  &MAE/MSE/NAE  $\downarrow$ & Avg. &$A0 \sim A5$ $\uparrow$ \\
		\whline
		Faster RCNN \cite{ren2015faster}  & ResNet-101  & 6.7/\textbf{95.8}/3.5 &414.2/1063.7/0.791 &18.2 &0/0.002/0.4/7.9/37.2/63.5  \\
		\hline
		TinyFaces \cite{hu2017finding}  &ResNet-101   &56.7/52.9/61.1  &272.4/764.9/0.750  & 59.8 & 4.2/22.6/59.1/\textbf{90.0}/\textbf{93.1}/\textbf{89.6}    \\
		\hline
		VGG+GPR \cite{gao2019c,gao2019domain}  &VGG-16  & 52.5/55.8/49.6  &127.3/439.9/0.410  & 37.4 & 3.1/27.2/49.1/68.7/49.8/26.3 \\
		\hline
		RAZ\_Loc \cite{liu2019recurrent} &VGG-16  &59.8/66.6/54.3 &151.5/634.7/0.305 &42.4 & 5.1/28.2/52.0/79.7/64.3/25.1   \\
	   \hline
        Crowd-SDNet \cite{wang2021self} &ResNet-50 &63.7/65.1/62.4 	&-/-/- &55.1& 7.3/43.7/62.4/75.7/71.2/70.2\\
        \hline
		PDRNet*  &Unknown  & 65.3/67.5/63.3  & 89.7/\textbf{348.9}/0.261  & 47.0 & 7.4/38.9/63.2/82.9/65.0/24.6 \\
		\hline
        TopoCount \cite{abousamra2020localization} &VGG-16  & 69.2/68.3/70.1  &107.8/438.5/-   & \textbf{63.3} & 5.7/39.1/72.2/85.7/87.3/89.7 \\
        \hline
		RDTM \cite{liang2021reciprocal} &VGG-16  & 69.9/75.1/65.4  &97.3/417.7/0.237   & 45.7 & 11.5/46.3/68.5/74.9/54.6/18.2 \\
		\hline
		SCALNet \cite{cheng2021decoupled} &DLA-34 & 69.1/69.2/69.0  &85.5/361.5/0.221   &- &-\\
		\hline
		D2CNet \cite{cheng2021decoupled} &VGG-16  & 70.0/74.1/66.2  &86.8/339.9/0.218   &58.3 & 11.3/ 50.2/67.8/74.5/69.5/76.5 \\
		\hline
		\multirow{2}{*}{IIM \cite{gao2020learning}}  &VGG-16  &73.2/77.9/69.2 &96.1/414.4/0.235 &58.7 & 10.1/44.1/70.7/82.4/83.0/61.4   \\
		\cline{2-6}
		  &HRNet  &76.2/81.3/71.7 &87.1/406.2/\textbf{0.152} &61.3 & 12.0/46.0/73.2/85.5/86.7/64.3   \\
        \whline
        \multirow{2}{*}{LDC-Net }
		  &VGG-16  &74.2/79.3/69.7 &83.1/394.5/0.214 &58.3 & 9.4/42.6/70.9/81.4/83.0/62.4  \\
        \cline{2-6}
		  &HRNet  &\textbf{76.3}/78.5/\textbf{74.3} &\textbf{80.4}/360.4/0.208& 56.6 &\textbf{14.8}/\textbf{53.0}/\textbf{77.0}/85.2/70.8/39.0  \\
		\whline
	\end{tabular}
	\label{table:nwpu_loc_count}
\end{table*}
\begin{figure*}[htbp]
	\centering
	\includegraphics[width=1\textwidth]{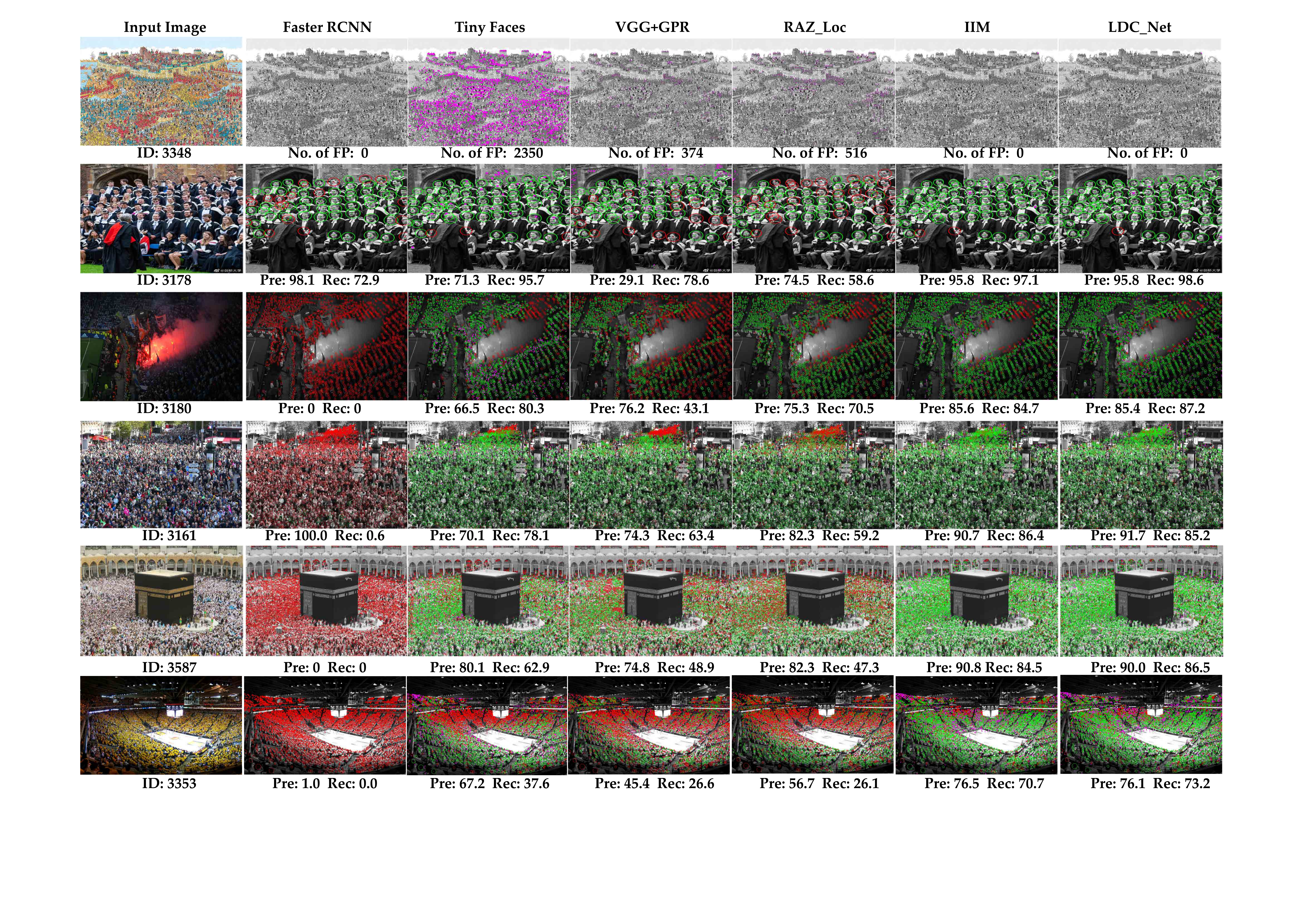}
	\caption{Some typical visualization results of four popular methods and the proposed IIM on NWPU-Crowd \emph{validation set}. The {\color{green}{green}}, {\color{red}{red}} and {\color{magenta}{magenta}} points denote true positive (TP),  false negative (FN) and  false positive (FP), respectively. The {\color{green}{green}} and {\color{red}{red}} circles are GroundTruth with the  radius of $\sigma_l$. For easier reading, images are transformed to gray-scale data.}
	\label{fig:vis_nwpu}
	\vspace{-0.2cm}
\end{figure*}
\begin{figure*}[htbp]
	\centering
	\includegraphics[width=1.\textwidth]{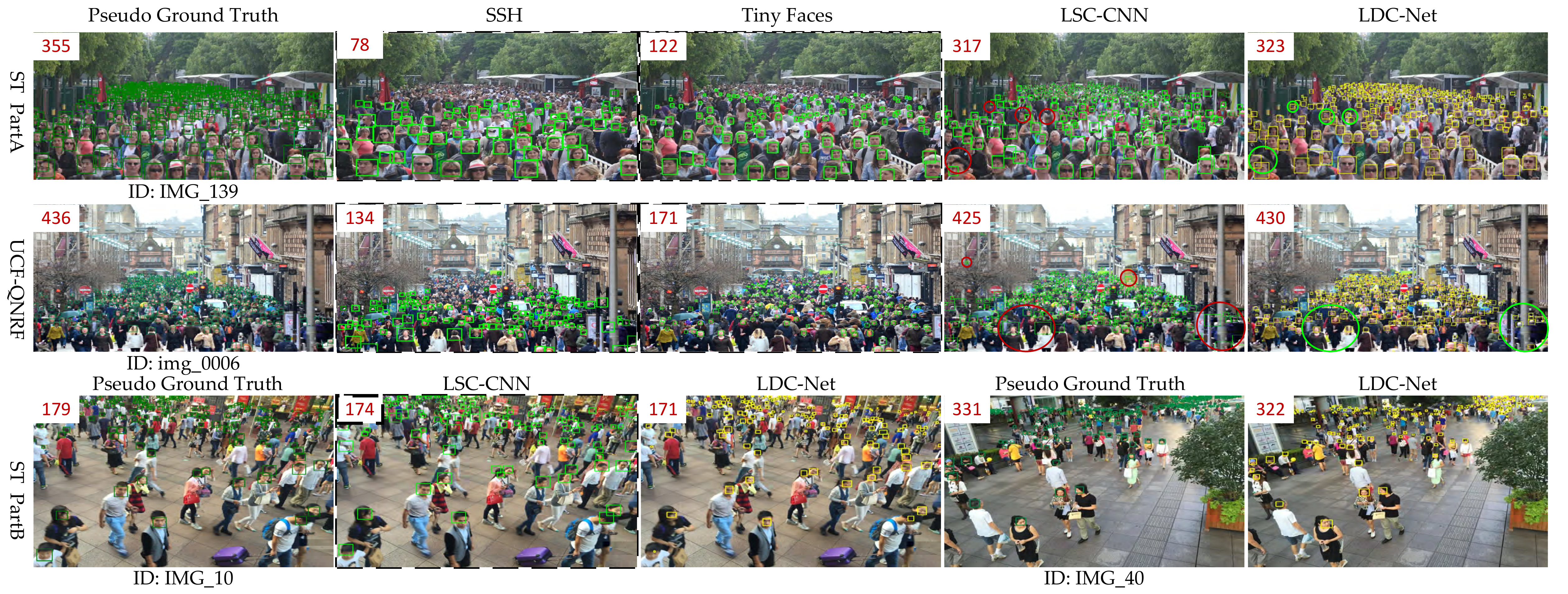}
	\caption{Visualization of the counting and detection performance on other three datasets (\emph{test set}). The images surrounded with dashed boxes are provided by \cite{sam2020locate}. {\color{red}{Red}} numbers represent the count. The {\color{yellow}{yellow}} boxes are our detection results, while the light {\color{green}{green}} boxes are other methods. Enlarge the images for better viewing.}
\label{Fig:vis_det_NWPU}
\end{figure*}
\subsubsection{Effect of the OUSR Loss}
\label{sec:ousr_loss}
 The differential binarization module makes it possible to design a task-oriented loss on the binary map, which usually brings a dominant performance improvement, such as the IOU loss in the semantic segmentation and object detection tasks. In this paper, we propose an OUSR loss to measure the quality of the binary map. In this part, we thoroughly discuss the contribution of OUSR Loss in what aspects. It takes two steps to explore the OUSR loss: 1) Firstly, we compare the experimental results that are with and without the OUSR loss, respectively; 2) Secondly, we take a roughly hyper-parameter search to confirm a better weight $\lambda$ in Eq. \ref{Eq:19}. As this part needs to train a model from the beginning with different $\lambda$, we arrange it on the ShangTech Part B \cite{zhang2016single} dataset due to the limited computation resource. The results are evaluated by the counting and localization tasks. Note that the ShanghaiTech dataset does not provide the box annotation, so we calculate the MLE and AP/AR/F1-score metrics to evaluate localization performance. As displayed in Table \ref{Table:OUSR_loss}, we also report the results of LSC-CNN \cite{sam2020locate} for a comparison. Lines 5, 6, and 7 are the results of LDC-Net with setting different $\lambda$ for OUSR loss, which surpass the LDC-Net in localization task and with a competitive counting performance.

From Table \ref{Table:OUSR_loss}, OUSR loss indeed contributes to the LDC-Net with a proper $\lambda$ by comparing Lines 5, 6, and 7. Specifically, OUSR loss help achieve better counting and localization performance through promoting the recall ratio as shown in the last row of Table \ref{Table:OUSR_loss} and Table \ref{Table:ablation}. It also verifies that the OUSR loss is viable in LDC-Net. Conspicuously, OUSR loss does not always have a positive effect. A large weight can also bring a side effect as line 6 shown. In Table \ref{Table:OUSR_loss}, $\lambda$ as 0.01 attains better results and we use this weight in subsequent experiments.

\subsection{Localization and Counting Performance}

\subsubsection{Performance on NWPU-Crowd}

The NWPU-Crowd provides a fair comparison environment by submitting the results for official evaluation. Here, LDC-Net is compared with the state-of-the-art methods on the NWPU-Crowd benchmark. Table \ref{table:nwpu_loc_count} shows the Overall performances (Localization: F1-m, Pre. and Rec.; Counting: MAE, MSE, and NAE) and per-class Rec. at the box level. By comparing the primary key (Overall F1-m and MAE), LDC-Net ranks first place (F1-m of 76.3\% and MAE of 80.4) in all crowd localization methods. Per-class Rec. on Box Level shows the sensitivity of the model for the instances with different scales. As the results in A0$\sim$A2 shown, LDC-Net performs better in catching small objects. In addition, TinyFaces performs well in A3$\sim$A5, achieving Rec. of $\sim$90\%. Compared with the IIM \cite{gao2020learning}, LDC-Net achieves a little promotion in overall localization performance and has a big progress in counting performance. By comparing the LDC-Net and other methods with the VGG-16 backbone, LDC-Net also surpasses other methods with a dominant gap.

To intuitively compare the localization results, we follow the visualization tools \cite{gao2020nwpu} to draw the visualization images for other methods and LDC-Net. Fig. \ref{fig:vis_nwpu} illustrates six groups of typical samples (negative sample, sparse, lower illumination and, extremely dense crowd scenes ) on NWPU-Crowd \emph{validation set}. In the first sample, LDC-Net shows a strong ability in avoiding false-negative predictions compared with some density-based localization methods (e.g., VGG+GPR and RAZ\_Loc). For a sparse scene (Row 2 in Fig. \ref{fig:vis_nwpu}), LDC-Net achieves the best results among all methods, which has one false positive and one false negative (the other two false positives in the middle are caused by the omissive annotations). In the dense crowd scenes (Id 3180, 3161, 3587, and 3353),  Faster R-CNN, a general object detection framework, loses almost all the heads. TinyFaces, VGG+GPR, and RAZ\_Loc perform better but output many false positives and false negatives. LDC-Net performs well when facing extremely congested scenes. Even in the lower illumination scene (Row 3 in Fig. \ref{fig:vis_nwpu}), LDC-Net can make promising location results. Noticeably, LDC-Net inherits the thoughts of IIM and improves its performance by significantly promoting the recall rate with little precision loss. Overall, LDC-Net is a dazzling localization framework that can handle diverse crowd scenes and show the robustness for background objects and negative samples. It can be further exploited in the crowd localization community as a solid baseline.

\begin{table*}[htbp]
	\centering
	\small
	\caption{The performance of some classical crowd localization methods and the proposed LDC-Net on four datasets.``*'': The results are provided by IIM \cite{gao2020learning}.}
	\begin{tabular}{cIcIc|c|cIc|c|cIc|c|cIc|c|c}
		\whline
		\multirow{2}{*}{Method} & \multirow{2}{*}{Backbone}&\multicolumn{3}{cI}{ShanghaiTech Part A} &\multicolumn{3}{cI}{ShanghaiTech Part B} &\multicolumn{3}{cI}{UCF-QNRF} &\multicolumn{3}{c}{FDST}\\
		\cline{3-14}
		&& \textbf{F1-m} & Pre. &Rec. &\textbf{F1-m} & Pre. &Rec. & \textbf{F1-m} & Pre. &Rec. &\textbf{ F1-m} & Pre. &Rec.\\
		\hline
		TinyFaces* \cite{hu2017finding} &ResNet-101 &57.3 &43.1 &\textbf{85.5} &71.1 &64.7 &79.0 &49.4 &36.3 &\textbf{77.3} &85.8 &86.1 &85.4  \\
		\hline
		RAZ\_Loc* \cite{liu2019recurrent} &VGG-16 &69.2 &61.3 &79.5 &68.0  &60.0 &78.3 &53.3 &59.4  &48.3 &83.7 &74.4&95.8  \\
		\hline	
		LSC-CNN* \cite{sam2020locate} &VGG-16 &68.0&69.6 &66.5 &71.2 &71.7 &70.6 &58.2 &58.6 &57.7 &- &- &-  \\
		\hline
		\multirow{2}{*}{IIM \cite{gao2020learning}}&VGG-16 &72.5  &72.6  &72.5   &80.2   & 84.9  &76.0   &68.8   &78.2   &61.5   &93.1  &92.7  &93.5   \\
		\cline{2-14}
		&HRNet &73.9 &\textbf{79.8} &68.7 &\textbf{86.2} &\textbf{90.7} &82.1 &72.0 &\textbf{79.3} &65.9 &95.5 &95.3&95.8  \\
		\whline
		\multirow{2}{*}{LDC-Net} &VGG-16&73.2  &73.2  &73.2     &84.6 &84.4 &84.8    &72.8 &77.3 &68.9    &95.7 &95.8 &95.6 \\
        \cline{2-14}
         &HRNet   &\textbf{74.8} &75.7 &73.9     &85.3 &85.7 &\textbf{85.0}    &\textbf{74.3} &78.1 &70.9 &\textbf{96.0} &\textbf{95.9}&\textbf{96.1}  \\
        \whline	
\end{tabular}
	\label{table:loc_other}
	\vspace{-0.3cm}
\end{table*}

\begin{table*}[htbp]
    \small 	
	\centering
	\caption{The localization  accuracy on four datasets with the metric (AP/AR/F1-score under 100 distance thresholds) proposed in \cite{idrees2018composition} and (MLE) proposed in LSC-CNN \cite{sam2020locate}. ``*'': The results are calculated with their pre-trained model.}
	\setlength{\tabcolsep}{0.95mm}{\begin{tabular}{cIcIcIcIcIc|cIc|c}
			\whline
			\multirow{2}{*}{Method}  &\multicolumn{2}{cI}{Shanghai Tech Part A} &\multicolumn{2}{cI}{Shanghai Tech Part B} &\multicolumn{2}{cI}{UCF-QNRF}&\multicolumn{2}{c}{NWPU-Crowd \emph{val set}}\\
			\cline{2-9}
			&\textbf{MLE}&\textbf{F1-score}/AP/AR &\textbf{MLE }&\textbf{F1-score}/AP/AR &\textbf{MLE}&\textbf{F1-score}/AP/AR &\textbf{MLE }&\textbf{F1-score}/AP/AR	\\
			\whline
	LSC-CNN* \cite{sam2020locate}             &12.0 &81.7/83.6/79.8   &9.0 &86.7/87.3/86.1     &8.6 &74.1/74.6/73.5	             & - &-/-/-\\
	TopCount* \cite{abousamra2020localization} &10.3&\textbf{85.3}/\textbf{86.3}/84.3    &7.8&\textbf{91.5}/\textbf{91.7}/\textbf{91.3}                  &-&80.3/81.8/79.0	     & - &-/-/-\\
			\whline
	LDC-Net(VGG16+FPN) &10.0  &85.1/85.0/\textbf{85.1}    &7.7 &90.8/90.6/91.0   &8.5 &80.8/82.6/79.4   &9.9 &77.9/79.1/76.8	\\
            \hline
	LDC-Net(HRNet) &\textbf{10.0}  &84.9/85.8/84.1    &\textbf{7.6} &91.1/91.5/90.7   &\textbf{8.1} &\textbf{81.7}/\textbf{83.2}/\textbf{80.3}   &\textbf{9.3} &\textbf{80.9}/\textbf{82.0}/\textbf{79.9}	\\
			\whline			
	\end{tabular}}
	\label{table:loc_other_mle}
\end{table*}

\begin{table*}[htbp]
    \small
    \newcommand{\tabincell}[2]{\begin{tabular}{@{}#1@{}}#2\end{tabular}}
	\centering
	\caption{The Counting  Performance of classical crowd counting methods and the proposed LDC-Net on other datasets. ``-'': The authors do not provide the result. ``*'': The results are provided by \cite{sam2020locate}. ``D'', ``P'', and ``P$\&$B'' represent the method output density map, point, point $\&$ box for counting, respectively. }
	\setlength{\tabcolsep}{1.7mm}{\begin{tabular}{cIcIc|c|cIc|c|cIc|c|cIc|c|c}
		\whline
		\multirow{2}{*}{Method} & \multirow{2}{*}{\tabincell{c}{Output \\ Type}}&\multicolumn{3}{cI}{ShanghaiTech Part A} &\multicolumn{3}{cI}{ShanghaiTech Part B} &\multicolumn{3}{cI}{UCF-QNRF} &\multicolumn{3}{c}{FDST}\\
		\cline{3-14}
		&& MAE & MSE &NAE & MAE & MSE &NAE  & MAE & MSE &NAE  &  MAE & MSE &NAE \\
        \hline
         MCNN \cite{Zhang2016CVPR}&  D &110.2 &173.2&-   &26.4 &41.3 &-     &277.0 &426.0 &-   &-&-&- \\
         CSRNet \cite{li2018csrnet}&  D &68.2 &115.0 &-   &10.6 &16.0 &-     &- &- &-           &-   &- &- \\
         PSDNN+ \cite{liu2019point}&  D  &65.9 &112.3 &-   &9.1 &14.2  &-     &- &- &-           &- &- &- \\

		\whline
		RAZ\_Loc\cite{liu2019recurrent}   &P &71.6 &120.1 &-   &9.9  &15.6 &-  &135.0 &246.0  &-    &-&-&-  \\
		Idrees \cite{idrees2018composition}  &P  &- &- &-     &-  &-&-    &132.0 &191.0  &-           &-&-&-  \\
	    IIM  \cite{gao2020learning} &P  &69.3 &138.1 &-     &13.5  &28.1 &-    &142.6 &261.1  &-           &1.4&1.9&-  \\
	    TopoCount  \cite{abousamra2020localization} &P& \textbf{61.2} &104.6 &-  &7.8  &13.7 &-  &\textbf{89.0} &\textbf{159.0}  &-           &-&-&-  \\
		\whline
		Faster RCNN* \cite{ren2015faster}  &P$\&$B  &241.0 &431.6 &-     &-&-&-  &320.1 &697.6&-  &- &- &-  \\
		\hline
		SSH* \cite{najibi2017ssh}  &P$\&$B     &387.5 &513.4 &-   &-&-&-   &441.1 &796.6&-    &- &- &-  \\
		\hline
		TinyFaces* \cite{hu2017finding} &P$\&$B &237.8 &422.8 &-  &-&-&-      &336.8 &741.6&-     &- &- &-  \\
		\hline	
        PSDNN \cite{liu2019point}  &P$\&$B  &85.4 &159.2 &-     &16.1&27.9&-    &- &-&-           &- &- &-  \\
        \hline
		LSC-CNN \cite{sam2020locate}&P$\&$B &66.4 &117.0 &-   &8.1 &12.7 &-     &120.5 &218.2 &-     &- &- &-  \\
		\hline
        Crowd-SDNet \cite{wang2021self} &P$\&$B  &65.1 &\textbf{104.4} &-   &\textbf{7.8} &\textbf{12.6} &-   &- &- &-       &- &- &-  \\
        \whline
		LDC-Net(VGG16+FPN) & P$\&$B &67.1 &120.3 &0.148   &8.1 &13.5 &0.066   &124.8 &226.6 &0.184       &1.3 &1.9 &0.060\\
        LDC-Net(HRNet) & P$\&$B &66.4 &115.7 &\textbf{0.147}   &8.1 &14.6 &\textbf{0.062}   &115.7 &200.2 &\textbf{0.176} &\textbf{1.3} &\textbf{1.8} &\textbf{0.055}
  \\
		\whline
	\end{tabular}}
	\label{table:counting_other}
	\vspace{-0.3cm}
\end{table*}

\subsubsection{Performance on Other Datasets}
Except for NWPU-Crowd, we also test the LDC-Net on other common crowd datasets and compare the results with the SOTA methods. Table \ref{table:loc_other} and Table \ref{table:loc_other_mle} list the localization results and Table \ref{table:counting_other} reports the counting results.

\textbf{Localization.} Firstly, we compare LDC-Net with some typical open-sourced algorithms (TinyFaces \cite{hu2017finding}\footnote{ https://github.com/varunagrawal/tiny-faces-pytorch \label{tiny_face}}, RAZ\_Loc \cite{liu2019recurrent}\footnote{https://github.com/gjy3035/NWPU-Crowd-Sample-Code-for-Localization}, LSC-CNN \cite{sam2020locate}\footnote{https://github.com/val-iisc/lsc-cnn \label{lsc-cnn}}), and TopoCount \cite{abousamra2020localization}\footnote{https://github.com/ShahiraAbousamra/TopoCount \label{TopoCount}} with the evaluation protocol proposed in NWPU-Crowd. As those metrics need box-level annotation to calculate the distance threshold $\sigma_{l}$, we use the pseudo box annotation generated by IIM \cite{gao2020learning} for the point annotation datasets, Shanghai Tech and UCF-QNRF. The results in Table \ref{table:loc_other} show LDC-Net attains the best trade-off between the precision and recall on the four datasets. Especially when compared to similar method IIM, LDC-Net achieves a big promotion on the VGG-16 backbone, such as increasing the F1-measure by 5.1\% on Part B. Besides, The VGG-backbone and HRNet have similar performance, which reveals that  LDC-Net does not depend much on the backbone. In the two dense crowd datasets (ShanghaiTech A and UCF-QNRF), TinyFaces have the best recall rate by sacrificing accuracy. By comparing the detailed performance, TinyFaces and RAZ\_loc tend to produce higher recall but lower precision; LSC-CNN brings the most balanced Recall and Precision. Compared with IIM, LDC-Net has a similar Precision while producing a higher Recall. In general, the proposed method outperforms other state-of-the-art methods and further improves IIM's performance on the other datasets. Take Rec. as an example, LDC-Net with HR-Net increases 6.2\% recall rate averaged on Shanghai Tech and UCF-QNRF datasets.

Secondly, to further demonstrate the superiority of LDC-Net in the localization task, we also adopt the point-based metric to evaluate it on the Shanghai Tech and UCF-QNRF datasets. Table \ref{table:loc_other_mle} reports the results of LDC-Net and the latest SOTA methods in comparison of MLE and F1-score/AP/AR, where MLE represents the average localization error, and LDC-Net achieves the best MLE over all datasets.  Besides, it also has competitive performances in the F1-score/AP/AR metrics.
\begin{table*}[htbp]
	\centering
	\caption{The detection performance on the NWPU-Crowd (\emph{val set}). $A0 \sim A5$ respectively means that the head area is in  $[10^0,10^1]$, $(10^1,10^2]$,  $(10^2,10^3]$,  $(10^3,10^4]$,  $(10^4,10^5]$, and $>10^5$. The footnote represents the results are produced with the corresponding code.}
	\begin{tabular}{cIcIc|cIcIc}
		\whline
		\multirow{2}{*}{Method}	&\multirow{2}{*}{Backbone}  &\multicolumn{2}{cI}{F1-m/Pre/Rec(\%)}  &Rec at $A0 \sim A5$  & AP(\%)  \\
		\cline{3-6}
		&  & iou$\geq$0.3 & iou$\geq$0.5 &  iou$\geq$0.5  & iou$\geq$0.5  \\
		\whline
		Faster RCNN \cite{ren2015faster} \textsuperscript{\ref{faster-rcnn}}  & ResNet-101  &10.1 /78.4/5.4  &7.4/58.0/4.0   &0 /0/ 0.07/ 9.8/42.5/61.7   &3.4 \\
		\hline
        TinyFaces \cite{hu2017finding} \textsuperscript{\ref{tiny_face}} &ResNet-101   & 22.7/13.8/64.6      &19.2/11.6/54.7   &0/15.8/52.1/\textbf{85.9} /\textbf{85.4}/\textbf{80.0}  &24.5\\
		\whline

	\multirow{2}{*}{LDC-Net}  &VGG-16  &61.7/60.8/62.7 &40.0/39.5/40.5  & 3.1/44.1/57.2/62.4/43.0/32.8 &28.7   \\
		\cline{2-6}
    &HRNet  &\textbf{68.2}/\textbf{68.5}/\textbf{71.9}  &\textbf{44.7}/\textbf{43.8}/\textbf{45.7} & \textbf{4.5}/\textbf{48.2}/\textbf{67.3}/65.0/48.1/30.0 &\textbf{32.4}    \\
		\whline
	\end{tabular}
	\label{table:nwpu_det}
\end{table*}
\begin{table*}[htbp]
	\centering
	\caption{The detection performance on the FDST (\emph{test set}). $A1 \sim A4$ respectively means that the head area is in  $(10^1,10^2]$,  $(10^2,10^3]$,  $(10^3,10^4]$, and $(10^4,10^5]$. The footnote represents the results are produced with the corresponding code.}
	\begin{tabular}{cIcIc|cIc|c}
		\whline
		\multirow{2}{*}{Method}	&\multirow{2}{*}{Backbone}  &\multicolumn{2}{cI}{F1-m/Pre/Rec(\%)}  &Rec at $A1 \sim A4$  & AP(\%)  \\
		\cline{3-6}
		&  & iou$\geq$0.3 & iou$\geq$0.5 &  iou$\geq$0.5  & iou$\geq$0.5  \\
		\whline
		TinyFaces \cite{hu2017finding} \textsuperscript{\ref{tiny_face}} &ResNet-101   & 85.3/84.9/85.7  &78.5/78.1/78.8   &42.5 / 72.3/ 87.4/ 97.1 &70.3\\

		\whline
		\multirow{2}{*}{LDC-Net}  &VGG-16  &91.2/92.4/92.1       &86.1/85.6/86.7       & 68.4/83.3/92.7/82.3                 &84.8   \\
		\cline{2-6}

		 &HRNet  &\textbf{94.1}/\textbf{93.8}/\textbf{94.3}      &\textbf{89.3}/\textbf{89.1}/\textbf{89.6}     & \textbf{70.0}/\textbf{85.5}/\textbf{94.9}/\textbf{85.7}  &\textbf{87.9}    \\
		\whline
	\end{tabular}
	\label{table:FDST_det}
\end{table*}

\textbf{Counting.}\quad In addition to localization, LDC-Net can estimate the number of people as well as some density-based methods do. In Table \ref{table:counting_other}, we compare LDC-Net's counting metrics in terms of MAE, MSE, and NAE against other SOTA methods. The current counting methods are divided into three categories according to the output formats, which are Density-based (D), Point-based (P), and Point$\&$Box-based (P$\&$B) methods. For the past few years, density-based methods perform as a navigator in the counting field. It can achieve charming counting results but has a limitation on location prediction. Hence, some methods (\cite{ren2015faster,najibi2017ssh,hu2017finding}) directly output point or box for counting. However, it has a deficiency in counting performance, which is usually hard to detect very small or large targets by suffering from the long-tail distribution. LDC-Net also sums the detected boxes or locations as a counting value.  Overall, LDC-Net achieves competitive performance against the SOTA counting methods, especially compared with the similar method, IIM. It is noticeable that some very tiny heads maybe only occupy few pixels in our label, which is very difficult to overtake with the segmentation. It is the deficiency of the LDC-Net in counting compared with the density-based methods. How to further improve its counting performance in this situation can be a research point in the future.

To visually compare the counting performance of LDC-Net with other methods, we draw a visual comparison between LDC-Net and several methods (SSH \cite{najibi2017ssh}, TinyFaces \cite{hu2017finding}, and LSC-CNN \cite{sam2020locate}). The samples are selected from those used in LSC-CNN. In terms of counting, the performance of LDC-Net is better than the others in dense scenarios. As shown in Row 1, LSC-CNN column, the circled head in red is the omitted count, but in LCD-Net, they are not missing. Furthermore, the box prediction in LDC-Net looks more reasonable and accurate, such as the red and green circles in Row 2.

\begin{figure*}[htbp]
	\centering
	\includegraphics[width=0.99\textwidth]{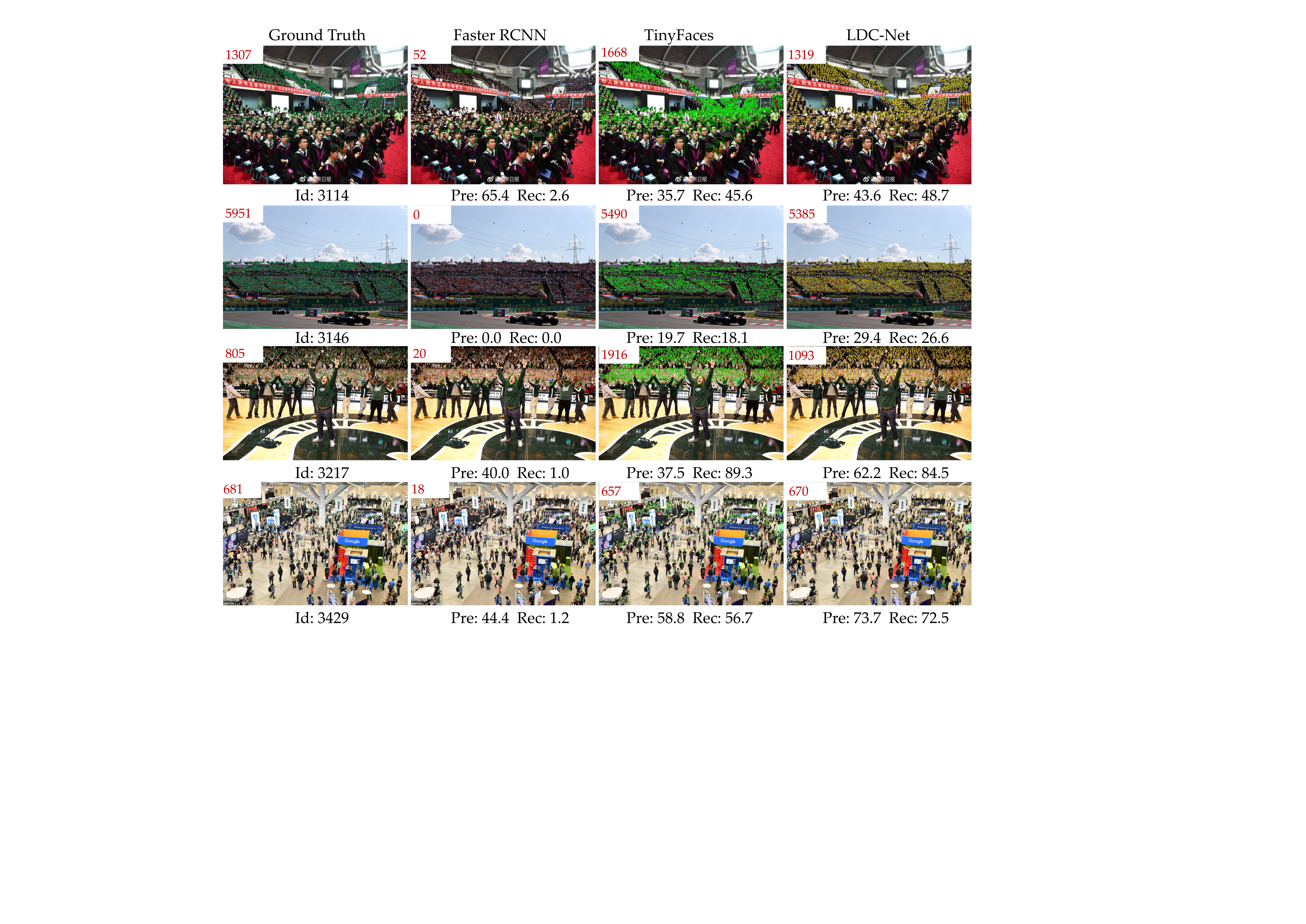}
	\caption{Visualization of the detection performance on NWPU-Crowd \emph{val set}. Pre and Rec scores are computed with the IOU threshold as 0.5. {\color{red}{Red}} numbers represent the count of the bounding box. Enlarge the images for better viewing.}
\label{Fig:vis_det_NWPU}
\end{figure*}

\subsection{Detection Performance}
Another feature of LCD-NET is competent to the detection task in the dense crowds. Here, we discuss the detection performance on three benchmark datasets, NWPU-Crowd \cite{gao2020nwpu}, FDST \cite{fang2019locality}, and WIDE RFACE \cite{yang2016wider}.

\textbf{Detection on NWPU-Crowd.}\quad NWPU-Crowd is a challenging dataset with a large-scale span in head size, usually ranging from 0$\sim$100,000+ pixels. As the annotation information of the test set is not available, we make the metric comparison at its validation set. Table \ref{table:nwpu_det} shows the detection performance between LDC-Net and two typical detection algorithms, Faster RCNN \cite{ren2015faster} \footnote{https://github.com/ruotianluo/pytorch-faster-rcnn \label{faster-rcnn}}, and TinyFaces \cite{hu2017finding}\textsuperscript{\ref{tiny_face}}. Table \ref{table:nwpu_det} describes the detection performance from three aspects: 1) F1-m/pre/Rec scores with the IOU threshold as 0.3 and 0.5, respectively. we first conduct the one-to-one matching strategy to judge the TP, FP, and FN with the IOU threshold, where the computation process is similar to the localization task used in NWPU-Crowd. Finally, the scores are calculated over the whole validation set. 2) The recall rates for different size objects are measured with the IOU threshold as 0.5, which reflects the sensitivity of the detection methods. 3) The overall AP is a universal metric in the object detection field, and we also introduce it to evaluate the head detection task. In Table \ref{table:nwpu_det}, LDC-Net gets the  best results on all metrics. Take the AP as an example, LDC-Net with HR-Net promotes 32.2\% compared with TinyFaces.

 To visually compare the detection performance, Fig. \ref{Fig:vis_det_NWPU} depicts some detection samples in NWPU-Crowd validation set with the LDC-Net, Faster RCNN, and TinyFaces. Faster RCNN almost failed in all dense crowds. It verifies that the traditional object detection methods are hard to catch the small size head. TinyFace is designed to detect the face, which has some similarities with head detection. TinyFaces get a good result compared with Faster RCNN, but the downside is that it produces many proposals and causes a lower precision. On the contrary, LDC-Net achieves higher precision as barely redundant proposals are generated. Besides, the predicted box size and the actual head size seem to fit better in LDC-Net, which also verifies our size prediction strategy is more accurate and effective.

\begin{figure}[t]
	\centering
	\includegraphics[width=0.4\textwidth]{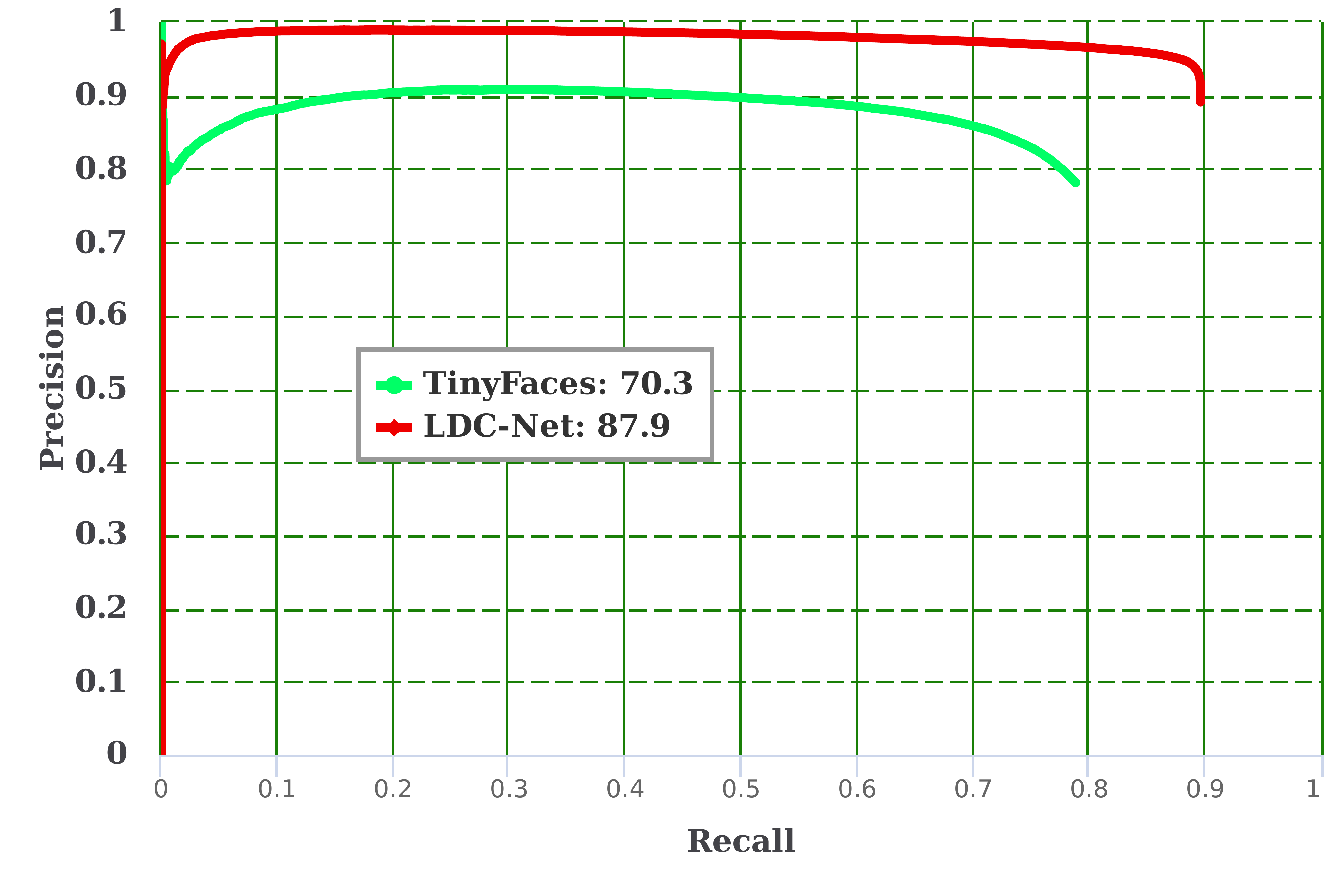}
	\caption{Precision and recall curves on FDST \emph{test set}.}
\label{Fig:pr_FDST}
\end{figure}
\textbf{Detection on FDST.}\quad FDST is a larger dataset with the box annotation, which fits the detection task well. To adopt the same metric for evaluation as NWPU-Crowd, we divide the head size from A1$\sim$A4 (10 pixel-10,000+ pixel ). Table \ref{table:FDST_det} shows the detection performance between TinyFaces and LDC-Net, LDC-Net wins in all metrics with high marks. Even with the VGG-16 backbone, it still achieves more accurate detection. LDC-Net produces 87.9 AP with the HRNet, which increases by 25\% compared with TinyFaces. Fig. \ref{Fig:pr_FDST} draws the precision-recall curves of two methods, it can be seen that LDC-Net has the largest envelope area and does not lose too much precision with the increase of recall ratio. This shows that LDC-Net can achieve a good balance between precision and recall.
\begin{table}[htbp]
	\centering
    \small
	\caption{The detection performance on the WIDER FACE (\emph{val set}), which evaluates the AP with the IOU threshold as 0.5.}
	\begin{tabular}{cIcIc|c|c}
		\whline
		\multirow{2}{*}{Method}&\multirow{2}{*}{Type}&\multicolumn{3}{c}{AP$_{0.5}$(\%)}  \\
		\cline{3-5}
	     && Easy& Medium &Hard  \\
     	\whline
	     ACF-WIDER  \cite{yang2014aggregate}&\multirow{7}{*}{Face} & 65.9&  54.1 & 27.3\\

		Faceness \cite{yang2015facial} & &80.8&  77.4 & 64.1  \\
	
		Two Stage CNN \cite{yang2016wider}&  &68.1&  61.8 &  32.3 \\
	
		LDCF+ \cite{ohn2016boost} & &79.0& 76.9 &  52.2 \\
	
		Faster RCNN \cite{ren2015faster} & &84.0 &72.4  &34.7 \\
		TinyFaces \cite{hu2017finding}&  &92.5&   91.0&  80.6  \\

		SSH \cite{najibi2017ssh} & & 93.1  &  92.1 & 84.5 \\
		\whline
		LSC-CNN \cite{sam2020locate} &\multirow{2}{*}{Head} & 57.3  &  70.1 & 68.9 \\

        LDC-Net     & &86.0 &85.9 &75.9    \\
		\whline
	\end{tabular}
	\label{table:WIDER}
\end{table}

\begin{figure*}[t]
	\centering
	\includegraphics[width=1.0\textwidth]{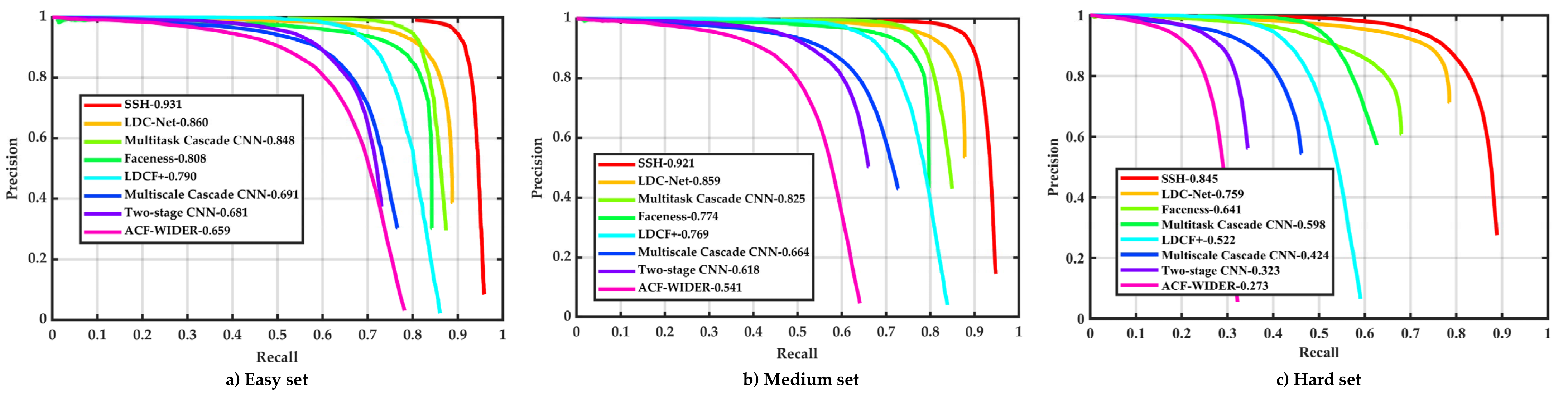}
	\caption{Precision and recall curves of different subsets on WIDER FACE \emph{val set}.}
\label{Fig:pr_WIDER}
\end{figure*}
\textbf{Detection on WIDER FACE.}\quad Table \ref{table:WIDER} lists the AP results of LDC-Net and other typical face detection methods. Our score is obtained by transferring LDC-Net to perform once training and then testing on WIDER FACE \emph{val set}. LDC-Net does not surpass some well-performed face detectors, such as TinyFaces \cite{hu2017finding} and SSH \cite{najibi2017ssh}. But it performs better than some two-stage object detection methods, such as Two-stage CNN \cite{yang2016wider} and Faster RCNN \cite{ren2015faster}. It also demonstrates that LDC-Net can be adapted to the face detection task and has a tremendous potential to be tapped. Besides, compared with the same type of crowd detection method LSC-CNN \cite{sam2020locate}, LDC-Net is leading with a huge margin. Fig. \ref{Fig:pr_WIDER} shows the performance curves of LDC-Net and other methods in three subsets. Although LDC-Net is not the best performance curve, we can also find that it has an advantage in maintaining the highest accuracy with increasing the recall.

\subsection{Computational Statistics}
Table \ref{Table:GFLOPs} shows the computation statistics in the LSC-CNN \cite{sam2020locate}, TopoCount\cite{abousamra2020localization}, and LDC-Net, including parameter size, Multiply-Accumulate Operations (MACs), forward time per image, and post-processing time per image. MACs measure the computation complexity of the model, and we test it by inputting the image with a size of 768$\times$1024. Forward time refers to the period taken from the inputting image to output the binary map (heat map in LSC-CNN). Post-processing refers to the time interval of getting the head location and bounding box from a binary map or heat map. Post-processing in LDC-Net and TopoCount are done on the CPU, while LSC-CNN is with the GPU. Following the LSC-CNN, the two times are averaged over the Shanghai Tech Part B test set. The GPU device is NVIDIA TITAN RTX (24G) and the CPU is Intel Xeon Silver 4110@2.10GHz. The statistics for LSC-CNN \textsuperscript{\ref{lsc-cnn}} and TopoCount \textsuperscript{\ref{TopoCount}} are tested with their realeased code .

In Table \ref{Table:GFLOPs}, LDC-Net has smaller MACs and faster inference time compared with LSC-CNN and TopoCount. Overall, LDC-Net has the potential to be a real-time system by further optimizing the baseline. Particularly, the post-processing is implemented on the CPU and it takes a lot of time. How to further optimize post-processing time is the key to improving the overall speed.

\begin{table}[htbp]
	\centering
	\small
	\caption{The computational statistics comparison between the LSC-CNN \cite{sam2020locate}, TopoCount \cite{abousamra2020localization} and LDC-Net.}
	\setlength{\tabcolsep}{2pt}{
		\begin{tabular}{c|c|c|c|c}
			\whline
			\multirow{2}{*}{Method}	& \multirow{2}{*}{LSC-CNN}&\multirow{2}{*}{TopoCount }& \multicolumn{2}{c}{LDC-Net}\\
            \cline{4-5}
            && &HRNet&VGG16\\
            \hline
			params(MB)      &35.1& 25.8&67.7  &\textbf{20.7} \\
			\hline
			MACs (G)          &1247&797.2 &\textbf{372.7}  &412.4 \\
			\hline
			Forward Time (ms)   &4634 & 95 &150   &\textbf{80} \\
            \hline
            Post-processing (ms) &\textbf{4}&23 &33     &32 \\
            \whline
		\end{tabular}	
	}
\label{Table:GFLOPs}
\end{table}

\section{Conclusion}
\label{sec:Conclusion}
In this paper, we propose a unified baseline that can simultaneously address the crowd localization, detection, and counting tasks. For the localization and counting tasks, we adopt the independent instance maps as a medium to detect the head positions. With the proposed OUSR loss and the redesigned binarization module, we further optimize the IIM \cite{gao2020learning} to get a better location and counting performance. For detection in dense crowds, we propose a simple but creative method to model the head scale, which makes the detection in a thousand people scene becomes easy and accurate. The experiments over several challenging datasets show the LDC-Net has predominant performance in locating and detecting dense crowds. Besides, it has a competitive performance when applied to counting tasks. Overall, LDC-Net can serve as a potential baseline to help researchers understand the dense crowds from some aspects. In particular, it has a lot of space to be explored in counting tasks and improving detection accuracy. In addition, our research can also inspire crowd tracking, video-level crowd counting in dense crowds.

\ifCLASSOPTIONcaptionsoff
  \newpage
\fi



%

\bibliographystyle{IEEEtran}
\bibliography{IEEEabrv,reference}

%
%

%



\end{document}